# Constraint-Handling Techniques for Particle Swarm Optimization Algorithms

Mauro S. Innocente [*] and Johann Sienz [†]

*ADOPT Research Group, Civil and Computational Engineering Centre (C²EC), School of Engineering, Swansea University, Singleton Park, Swansea, SA2 8PP, UK*

**Population-based methods such as Evolutionary Algorithms (EAs) and Particle Swarm Optimization (PSO) have proven their ability to cope with a variety of remarkably different problems, regardless of whether they are or are not linear, convex, differentiable or smooth. In addition, they are able to handle problems of notably higher complexity than traditional methods. The main procedure consists of successively updating a population of candidate solutions, performing a parallel exploration instead of traditional sequential exploration (usually unable to overcome local pathologies). While the origins of the PSO method are linked to bird flock simulations, it is a stochastic optimization method in the sense that it relies on random coefficients to introduce creativity, and a bottom-up artificial intelligence-based approach in the sense that its intelligent behaviour emerges in a higher level than the individuals' rather than deterministically programmed. As opposed to EAs, the PSO involves no operator design and few coefficients to be tuned. Since this paper does not intend to study such tuning, general-purpose settings are taken from previous studies. The plain PSO algorithm is only able to deal with unconstrained problems, so that some technique needs to be incorporated to handle constraints. A popular one is the penalization method, which turns the original constrained problem into unconstrained by penalizing the function value associated to infeasible solutions. Other techniques can be specifically designed for PSO, such as the preserving feasibility and the bisection methods. Given that these strategies present advantages and disadvantages when compared to one another, there is no obvious best constraint-handling technique (CHT) for all problems, and modifications and new techniques are constantly proposed in the literature. The aim here is to develop and compare different CHTs suitable for PSOs, which are incorporated to an algorithm with general-purpose settings. The comparisons between the different CHTs are performed keeping the remaining features of the algorithm the same, while comparisons to other authors' results are offered as a frame of reference for the optimizer as a whole. Thus, three basic techniques: the penalization, preserving feasibility and bisection methods –as well as modifications that aim to overcome their weaknesses– are discussed, implemented, and tested on two suites of benchmark problems. Three neighbourhood sizes are also considered in the experiments.**

## Nomenclature

| | | |
|---|---|---|
| EA, GA, CDE | = | evolutionary algorithm, genetic algorithm, cultured differential evolution |
| SA, NSM, ES | = | simulated annealing, nonlinear simplex method, evolution strategy |
| SR | = | stochastic ranking |
| PSO, $nn$, $nrs$ | = | particle swarm optimization/optimizer, number of neighbours, number of runs |
| PESO, IPSO | = | particle evolutionary optimization, improved particle swarm optimization |
| AI, AL, CI, SI | = | artificial intelligence, artificial life, computational intelligence, swarm intelligence |
| CHT, $cv$, $nac$ | = | constraint-handling technique, constraints' violations, number of active constraints |
| $iw$, $sw$, $w$, $aw$ | = | individuality, sociality, inertia, and acceleration weights |
| $U_{(a,b)}$ | = | Random number from uniform distribution between "a" and "b" |
| $N_{(a,b)}$ | = | 0-mean random number from normal distribution and standard deviation equal to 1 |

Communicating Author: mauroinnocente@yahoo.com.ar
[*] Ph.D. student, mauroinnocente@yahoo.com.ar
[†] Professor & Programme Director for Mechanical and Aerospace Engineering, J.Sienz@swansea.ac.uk

| | | |
|---|---|---|
| $x_{ij}^{(t)}$, $v_{ij}^{(t)}$ | = | $j^{\text{th}}$ coordinate of the position and velocity, respectively, of particle $i$ at time-step $t$ |
| $pbest_{ij}^{(t)}$ | = | $j^{\text{th}}$ coordinate of best position found by particle $i$ by time-step $t$. |
| $lbest_{ij}^{(t)}$ | = | $j^{\text{th}}$ coordinate of best position found by any particle in $i^{\text{th}}$ particle's neighbourhood by time-step $t$ |
| PF, PFPR | = | preserving feasibility, preserving feasibility with priority rules |
| PFPPR | = | preserving feasibility with probabilistic priority rules |
| BM, BMEM | = | bisection method, bisection method with extra momentum |
| BMPEM, APM | = | bisection method with probabilistic extra momentum, additive penalization method |
| REC, *fes* | = | relaxed equality constraints, number of function evaluations |

## I. Introduction

The features of the variables and functions that model the problem severely limit the suitability of traditional algorithms. Typically, the objective variables, objective functions and constraint functions must comply with a number of requirements for a given traditional method to be applicable. Furthermore, traditional methods tend to converge towards local optima, as their sequential point-to-point search is commonly unable to overcome local pathologies. By contrast, population-based methods such as evolutionary algorithms (EAs) and particle swarm optimization (PSO) are general-purpose optimizers, which are able to handle different types of variables and functions with few or no adaptations. Therefore, they can cope with a variety of remarkably different problems such as optimization, data mining, pattern recognition, classification, machine learning, scheduling, decision making and supply-chain management, regardless of whether they are or are not continuous, linear, convex, differentiable or smooth. Note that these methods do not require auxiliary information such as gradients and Hessians, and only the objective function information is necessary to guide the search. In addition, although finding the global optimum is by no means guaranteed, they are able to escape poor local optima by performing a parallel search. The latter is carried out by a population of interacting individuals that profit from information acquired through experience, and use stochastic weights or operators to introduce new responses. The lack of limitations to the features of the variables and functions enable these methods to deal with problems whose high complexity does not allow traditional, deterministic approaches. Regarding their pitfalls, both PSO and EAs require relatively high computational effort, some constraint-handling technique (CHT) incorporated, and typically find it hard to cope with equality constraints. In turn, the main advantages of PSO when compared to EAs are its lower computational cost, easier implementation, fewer coefficients to be tuned, and lack of operators' design.

Although the PSO approach required the tuning of a few coefficients only, even marginal variations in their settings greatly influence the dynamics of the swarm. Since this paper does not intend to study the tuning of these coefficients, general-purpose settings are taken from previous studies (refer to Innocente[15] and Innocente et al.[16]). Nevertheless, a few modifications are derived from the experimental results, and improvements are proposed and tested.

In its purest version, PSO is suitable for continuous and unconstrained problems. While some adaptations are necessary to deal with discrete problems, CHTs need to be incorporated to handle constrained problems. Some techniques can be imported from the EAs literature due to the similarities between the approaches, keeping in mind that the PSO method is elitist. A popular technique is the *penalization method*, which transforms the original constrained problem into unconstrained by penalizing the objective function associated to infeasible solutions. Some other techniques can be specifically designed for PSO, such as the *preserving feasibility strategy* initially proposed by Hu et al.[17], the *reflection method* proposed by Foryś et al.[19], or the *bisection method* proposed by Innocente[15]. Each of these strategies presents advantages and disadvantages when compared to one another, so that there is no obvious best CHT for all problems. Hence modifications and new methods can be constantly found in the literature. The aim in this paper is to develop and compare different CHTs suitable for PSO, which are incorporated to an algorithm with general-purpose settings of its coefficients. The comparisons between the different CHTs implemented are performed keeping the remaining features of the algorithm the same, while the comparisons to other authors' results are offered as a frame of reference for the optimizer as a whole. Therefore, two suites of constrained benchmark problems are used in our experiments: the first one taken from Hu et al.[17] (one function added), and the second one taken from Toscano Pulido et al.[26]. Three techniques previously discussed and tested by Innocente[15], namely the *penalization*, *preserving feasibility* and *bisection* methods, as well as modifications to them that aim to overcome their weaknesses, are discussed, implemented, and tested. Three neighbourhoods are also considered in the experiments, consisting of a ring topology (three-particle neighbourhood), eleven-particle, and swarm-size neighbourhoods. Finally, a dynamic neighbourhood and few minor modifications to the general-purpose settings are proposed together with dynamic constraint-violation tolerances, and the improved algorithm is tested on the same suites of benchmark problems. Results are compared to those obtained by numerous different optimizers.

## II. Optimization

For problems where the quality of a solution can be quantified in a numerical value, *optimization* is the process of seeking the permitted combination of variables that optimizes that value. Thus, different combinations of *variables* allow trying different candidate solutions, the *constraints* limit the valid combinations, and the *optimality criterion* allows differentiating better from worse. When formulating a problem, the question is whether to introduce numerous simplifications so that the available methods are able to solve the model, or to develop a model with higher fidelity and approximate the solving techniques. The suitability of traditional (i.e. deterministic) methods is limited by the nature of the variables, and by requirements that the objective and constraint functions must comply with. This sometimes leads to models based on strong simplifying assumptions so as to suit the solver, resulting in an exact solution of an approximate model. The alternative is to develop more precise models, and then attempt to solve those using approximate techniques and modern heuristics, which results in an approximate solution of a relatively precise model. The second approach is usually better for complex, real-world problems.

We are only concerned with single-objective optimization problems within this work. While the *objective* of a problem is given in plain words, its *formulation* is called the *objective function*. The output of this function is some measure of the fulfilment of the objective. Thus, the *objective function* relates the real problem to the model, while the *cost function* –also *evaluation, fitness,* or *conflict function*– is the scalar function to be optimized. The objective and cost functions might coincide, or there might be some mapping between them. The *problem variables* are also called *object* or *design variables*, or just *variables*. Since *parameters* might stand for either *variables* or *coefficients*, the use of the term is avoided within this work.

The important differences among different types of optimization problems make it necessary to handle each type by means of completely different approaches. Optimization methods can be classified according to the *type of problems* they are able to handle (i.e. *linear* or *nonlinear*; *continuous, discrete* or *mixed-integer*; *convex* or *non-convex*; *differentiable* or *non-differentiable*; *smooth* or *non-smooth*; *etc.*), or according to the *features of the algorithms* (i.e. *gradient-based* or *gradient-free*; *exact* or *approximate*; *deterministic* or *probabilistic*; *analytical, numerical* or *heuristic*, *single-based* or *population-based*, etc.). *Heuristics* in this context refers to techniques that do not guarantee to find anything, and are usually based on common sense, natural metaphors, or even methods whose behaviour is not fully understood. As an optimization technique, the PSO method is a *gradient-free, approximate, probabilistic, heuristic, population-based* method, which can handle *continuous* problems regardless of whether they are or are not *linear, convex, differentiable* or *smooth*. *Discrete* or *mixed-integer* variables can be handled to some extent with some modifications to the basic algorithm, or by its binary version (beyond the scope of this paper).

### A. General continuous optimization problem

Let $S$ be the search-space, and $\mathcal{F} \subseteq S$ its feasible part. A minimization problem consists of finding $\hat{\mathbf{x}} \in \mathcal{F}$ such that $f(\hat{\mathbf{x}}) \leq f(\mathbf{x}) \; \forall \mathbf{x} \in \mathcal{F}$, where $f(\hat{\mathbf{x}})$ is a global minimum and $\hat{\mathbf{x}}$ its location.

The problem can be formulated as in Eq. (1):

$$\text{Minimize } f(\mathbf{x}), \quad \text{subject to } \begin{cases} g_j(\mathbf{x}) \leq 0 & ; \quad j = 1, \ldots, q \\ g_j(\mathbf{x}) = 0 & ; \quad j = q+1, \ldots, m \\ l_i \leq x_i \leq u_i & ; \quad i = 1, \ldots, n \end{cases} \quad (1)$$

where $\mathbf{x} \in S \subseteq \mathcal{R}^n$ is the vector of variables; $f(\cdot) : S \to \mathcal{E} \subseteq \mathcal{R}$ is the cost function; and $g_j(\cdot) : S \to \mathcal{G} \subseteq \mathcal{R}$ is the $j^{\text{th}}$ constraint function. Since $\max(f(\mathbf{x})) = -\min(-f(\mathbf{x}))$, optimization stands for minimization within this work, and maximization benchmark problems are conveniently reformulated. Similarly, the inequality constraints $g_j(\mathbf{x}) \leq 0$ are not restrictive, given that $g_j(\mathbf{x}) \geq 0$ can be reformulated as $-g_j(\mathbf{x}) \leq 0$. Besides, considering that $l_i \leq x_i \leq u_i$ is equivalent to $(x_i - u_i \leq 0 \; \wedge \; -x_i + l_i \leq 0)$ and $g_j(\mathbf{x}) = 0$ is equivalent to $(g_j(\mathbf{x}) \leq 0 \; \wedge \; -g_j(\mathbf{x}) \leq 0)$, the problem can be reformulated as in Eq. (2), without loss of generality.

$$\text{Minimize } f(\mathbf{x}), \quad \text{subject to } g_j(\mathbf{x}) \leq 0 \quad ; \quad j = 1, \ldots, q + 2 \cdot (m - q) + 2 \cdot n \quad (2)$$

Although constraints allow concentrating the search into limited areas, each potential solution must be verified to comply with all of them. Hence the problem usually becomes harder to solve than its unconstrained counterpart.

**B. Types of constraints**

The most appropriate technique to handle a constraint usually depends on the type of constraint:
1) *Inequality constraint*: function of the object variables that must be smaller than or equal to a constant, as shown in Eqs. (1) and (2).
2) *Equality constraint*: function of the object variables that must be equal to a constant, as shown in Eq. (1).
3) *Boundary constraint*: instance of inequality constraints, consisting of functions that define boundaries that contain the feasible space; if the boundary constraints are given by a hyper-rectangle, they are also called *interval* or *side constraints*.

While a constrained optimization problem was defined as the problem of finding the combination of variables that minimizes the cost function while satisfying all constraints, real-world problems sometimes do not lend themselves to such strict conditions. Frequently, all the constraints cannot be strictly satisfied simultaneously, and the problem turns into finding a trade-off between minimizing the cost function and minimizing the constraints' violations. A constraint is said to be *hard* if it does not admit any degree of violation, and *soft* if there is some given tolerance. Since a tolerance is required for PSO algorithms to cope with equality constraints, only soft constraints are considered within this work. In other words, hard equality constraints cannot be handled.

### III. Particle Swarm Optimization

The method was invented by social-psychologist J. Kennedy and electrical-engineer R C. Eberhart in 1995[8], inspired by earlier bird flock simulations framed within the field of social psychology. In particular, Reynolds' boids[9] and Heppner and Grenander's artificial birds[10] strongly influenced the early developments. Hence the method is closely related to other simulations of social processes and experimental studies in social psychology, while also having strong roots in optimization and artificial intelligence (AI).

From the *social psychology* viewpoint, it performs some simulation of a social milieu. Some experimental studies in social psychology that influenced the method are Lewin's field theory; Gestalt theory; Sherif's and Asch's experiments; Latané's social impact theory; Bandura's no-trial learning; simulations of spread of culture in a population, and simulations of the behaviour of social animals (refer to Kennedy et al.[7] or Innocente[15] for further reading).

From an optimization viewpoint, it is a search method suitable for optimization problems whose solutions can be represented as points in an *n*-dimensional space. While the variables need to be real-valued in its original version, binary and other discrete versions have also been developed (e.g. Kennedy et al.[7], Kennedy et al.[12], Clerc[4], Mohan et al.[6]). Since the method is not designed to optimize a given problem but to carry out some procedures that are not directly related to the optimization problem, it is frequently referred to as a *modern heuristics*. Optimization occurs, nevertheless, without obvious links between the implemented technique and the resulting optimization process.

From the AI point of view, the PSO paradigm is viewed as belonging to different branches such as *Artificial Life* (AL), *Computational Intelligence* (CI), and in particular, *Swarm Intelligence* (SI). The latter is concerned with the study of the collective behaviour that emerges from decentralized and self-organized systems. SI is the property of a system composed of individual parts that interact locally with one another and with their environment, inducing the emergence of coherent global patterns without a sense of purpose or central control. Thus, the PSO paradigm is a bottom-up SI-based technique because its ability to optimize is an *emergent* property that is not specifically implemented in the code. The system's intelligent behaviour emerges in a higher level than the individuals', evolving intelligent solutions without using the programmers' expertise on the subject matter. It relies on random coefficients to introduce creativity into the system. Thus, the problem per se is not deterministically solved, but artificial-intelligent entities are programmed, which are expected to find a solution themselves.

Either *modern heuristics* or *SI-based optimizers*, the PSO algorithm is not designed to optimize but to perform a sort of simulation of a social milieu, where the ability of the population (*swarm*) to optimize its performance emerges from the cooperation among individuals (*particles*). While this makes it difficult to understand the way optimization is actually performed, it shows astonishing robustness in handling many kinds of complex problems that it was not specifically designed for. It presents the disadvantage that its theoretical bases are extremely difficult to be understood in a deterministic manner. Nonetheless, considerable theoretical work has been carried out on simplified versions of the algorithm, extrapolated to the full version, and finally supported by experimental results (e.g. Trelea[1], Clerc et al.[2], Ozcan et al.[5]). For a comprehensive review, refer to Kennedy et al.[7], Engelbrecht[14], and Clerc3.

The function to be minimized is called here the *conflict function* due to the social-psychology metaphor that inspired the method: That is, each individual seeks the minimization of the conflicts among its beliefs by using information gathered from its own and others' experiences. Individuals seek agreement by clustering in the space of beliefs, which is the result of imitating their most successful peer(s), thus becoming more similar to one another as the search progresses. Clustering is delayed by their own experiences, which each individual is reluctant to disregard.

**A. Unconstrained PSO**

While the emergent properties of the PSO algorithm result from local interactions among *particles* in a *swarm*, the behaviour of a single particle can be summarized in three sequential processes:

*Evaluation*. A particle evaluates its position in the environment, given by the associated value of the conflict function. Following the social psychology metaphor, this stands for the conflict among its current set of beliefs.

*Comparison*. Once the particle's position in the environment is evaluated, it is not straightforward to tell how good it is. Experiments and theories in social psychology suggest that humans judge themselves by comparing to others, thus telling better from worse rather than good from bad. Therefore, the particle compares the conflict among its current set of beliefs to those of its neighbours.

*Imitation*. The particle imitates those whose performances are superior in some sense. In the basic PSO algorithm, only the most successful neighbour is imitated.

These three processes are implemented within PSO, where the only sign of individual intelligence is a small memory. The basic update equations are as follows:

$$v_{ij}^{(t)} = w \cdot v_{ij}^{(t-1)} + iw \cdot U_{(0,1)} \cdot \left(pbest_{ij}^{(t-1)} - x_{ij}^{(t-1)}\right) + sw \cdot U_{(0,1)} \cdot \left(lbest_{ij}^{(t-1)} - x_{ij}^{(t-1)}\right) \tag{3}$$

$$x_{ij}^{(t)} = x_{ij}^{(t-1)} + v_{ij}^{(t)} \tag{4}$$

where $x_{ij}^{(t)}$ and $v_{ij}^{(t)}$ are the $j^{th}$ coordinate of the position and velocity, respectively, of particle $i$ at time-step $t$; $U_{(0,1)}$ is a random number from a uniform distribution in the range [0,1] resampled anew every time it is referenced; $w$, $iw$ and $sw$ are the inertia, individuality, and sociality weights, respectively; $pbest_{ij}^{(t)}$ and $lbest_{ij}^{(t)}$ are the $j^{th}$ coordinate of the best position found by particle $i$ and by any particle in its neighbourhood, respectively, by time-step $t$.

As shown in Eqs. (3) and (4), there are three coefficients in the basic algorithm which rule the dynamics of the swarm: the inertia ($w$), the individuality ($iw$), and the sociality ($sw$) weights. The $iw$ and the $sw$ are sometimes referred to as the learning weights, while their aggregation is called here the acceleration weight ($aw$).

Thus, the performance of a particle in its current position is evaluated in terms of the *conflict function*. In order to decide upon its next position, the particle compares its current conflict to those associated to its own and to its neighbours' best previous experiences. Finally, the particle imitates both best experiences to some extent.

The relative importance given to $iw$ and $sw$ leads to more self-confident or conformist behaviour, while the random weights introduce creativity: since they are resampled anew for each time-step, for each particle, for each dimension, and for each term in Eq. (3), the particles display uneven trajectories that allow better exploration. In addition, resampling them anew for the individuality and the sociality terms −together with setting $iw = sw$− leads to the particles alternating self-confident and conformist behaviour. Typically, $iw = sw = 2$ (i.e. $aw = 4$). Every particle also tends to keep its current velocity, where the strength of this tendency is governed by $w$. The relative importance between $w$ and $aw$ results –broadly speaking– in more explorative or more exploitative behaviour of the swarm.

Kennedy et al.[8] did not consider $w$ (i.e. $w = 1$) in their original algorithm, and suggested setting $iw = sw = 2$. However, the particles tended to diverge (*explosion*). It was found that if the components of the particles' velocities were bounded, the *explosion* could be controlled, and the particles ended up clustering around a solution. Later, Shi et al.[13] incorporated the inertia weight ($w$) to control the explosion, while Clerc et al.[2] studied the trajectory of a simplified one-particle system with stationary attractors and no random weights, developing a constriction factor which would both control the explosion and ensure convergence on a local optimum (of a single non-random particle). The results were extrapolated to the full PSO algorithm and successfully tested on benchmark problems. Innocente[15] carried out geometrical analyses of the average behaviour of a single particle with stationary attractors, showing that both $w = 1$ and $w = 0$ lead to average cyclic behaviour, while $w = 0.5$ favours fast convergence (for $iw = sw = 2$).

Different settings of the coefficients in Eq. (3) notably affect the behaviour of the swarm. For instance, the greater the acceleration weight the higher the degree of randomness, while the individuality/sociality and the inertia/acceleration ratios control the degree of exploration and exploitation. The topology of the neighbourhoods in the swarm also affects the exploration and exploitation abilities of the algorithm. Kennedy et al.[8] developed the original PSO with a fully connected social network (global PSO, where the number of neighbours is $nn = swarm\text{-}size – 1$), while Eberhart et al.[11] proposed the first local version ($nn < swarm\text{-}size – 1$). That is, the neighbourhood sizes are smaller than the swarm size, and they overlap. Three typical neighbourhood topologies are shown in Fig. 1.

The topology on the left is the original, *fully connected topology*, where every particle is connected to every other. That is, every particle is connected to *swarm-size – 1* neighbours. The topology in the centre is called *ring topology*, where every particle is connected to two neighbours, and every neighbourhood overlaps with the previous and with the next ones in two particles. Hence every particle belongs to three neighbourhoods. The rightmost topology is the *wheel topology*, where one particle is connected to all others, and all others with this particle only. Numberless neighbourhood topologies are possible, and the optimal design appears to be problem-dependent. Note that the neighbourhood is typically defined topologically –only once, at the beginning–, so that neighbouring particles are not necessarily near one another in the search-space. It is self-evident that the local versions are better suited to avoid premature convergence and escape poor local optima, at the expense of slower convergence.

**B. Constrained PSO**

The plain PSO algorithm lacks a mechanism to handle constraints. Some CHTs can be imported from the EAs' literature, and others can be developed by exploiting specific features of the PSO algorithm. Different classifications can be found in the literature, but it is not straightforward to decide which technique belongs to which group (especially within the PSO framework). For instance, Engelbrecht[14] argues that the *preserving feasibility* methods are those which ensure that adjustments to the particles do not violate any constraint, and defines the *repair methods* as methods that allow particles to move into infeasible space, but special operators are then applied to either bring the particle back or towards feasible space. These methods usually require initial feasible swarm. Thus, the *bisection method* proposed by Innocente[15] fits into the class of *preserving feasibility* methods as defined by Engelbrecht[14], and the *preserving feasibility* method proposed by Hu et al.[17] fits into the *repair methods* defined by Engelbrecht[14]. Therefore, we make our own classification here, which does not intend to be controversial or comprehensive. Emphasis is put on those techniques suitable for PSO algorithms that were investigated by the authors. Four main groups are considered: the *preserving feasibility techniques*, the *penalization methods*, the *repairing methods*, and *multiobjective-based techniques*. The latter are only mentioned and referred to other authors' work, as none of them were implemented in our optimizer at its current stage.

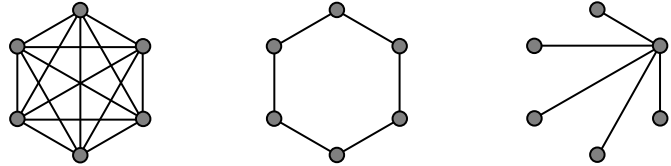

Figure 1. Three neighbourhood topologies. *Left*: fully connected topology ($nn$ = swarm-size – 1); *centre*: ring topology ($nn$ = 2); and *right*: wheel topology ($nn$ = swarm-size – 1 for one particle and $nn$ = 1 for the rest).

*1. Preserving feasibility techniques*

In agreement with Hu et al.[17], the preserving feasibility technique consists of the experience gathered by infeasible particles being ignored by the swarm and by the particle itself. Although the particles can fly over infeasible regions, they are quickly pulled back to feasible space as infeasible solutions are not stored in memory.

Its advantages are that a feasible solution is guaranteed, and that it only requires two small modifications to the unconstrained algorithm: successive random initialization until a feasible swarm is generated, and the feasibility condition for the update of the best experiences. Its drawbacks are that the random initialization might be extremely time consuming for low feasibility ratios of the search-space, and the lack of exploration of infeasible space. The latter might make it impossible to explore disjointed feasible spaces, for instance.

*2. Penalization methods*

The evaluation of infeasible solutions might be useful to guide the search towards more promising areas. Thus, the value of the *conflict function* is penalized for infeasible solutions (increased for minimization problems). Many different kinds of penalization methods can be found in the literature, according to the way the penalization is calculated. Thus, it can be linked to the number of constraints violated, although it is preferred to link it to the amount of constraints' violation. Multiplicative penalization methods, as opposed to the additive ones, also consider the value of the conflict function itself in the penalization. That is, the higher the conflict function value, the higher the penalization for the same degree of constraints' violation. We obtained better results with additive penalizations, although research on the subject was not extensive.

Penalization methods usually involve at least a couple of penalization coefficients that typically need to be tuned –which has been pointed as a drawback to the method–, although research on adaptive coefficients is extensive (e.g. Parsopoulos et al.[29], and Coello Coello[20]). With infeasible solutions penalized, the problem is treated as if it was unconstrained. A classical additive penalization (linked to the amount of infeasibility) is shown in Eqs. (5) and (6).

$$f_p(\mathbf{x}) = f(\mathbf{x}) + \sum_{j=1}^{m} \left[ k_j \cdot \left( f_j(\mathbf{x}) \right)^{\alpha_j} \right] \quad (5)$$

$$f_j(\mathbf{x}) = \begin{cases} \max\{0, g_j(\mathbf{x})\} & ; \quad 1 \leq j \leq q \\ \mathrm{abs}(g_j(\mathbf{x})) & ; \quad q < j \leq m \end{cases} \quad (6)$$

where $f(\mathbf{x})$ is the conflict function; $f_p(\mathbf{x})$ is the penalized conflict function; $f_j(\mathbf{x})$ is the amount of violation of $j^{th}$ constraint; and $k_j$ and $\alpha_j$ are penalization coefficients. The latter may be constant, time-varying, or adaptive, and they can be the same or different for different constraints. Typically, $k_j$ is set to high and $\alpha_j$ to small values. A high penalization might lead to infeasible regions of $S$ not being explored, converging to a non-optimal but feasible solution. A low penalization might lead to the system evolving solutions that are violating constraints but present themselves as having lower conflicts than feasible solutions. The definition of penalty functions is not trivial, and plays a critical role in the performance of the algorithm. Numerous penalization methods have been developed.

*3. Repairing techniques*

As mentioned before, Engelbrecht[14] defines them as methods that "…allow particles to move into infeasible space, but special operators (methods) are then applied to either change the particle into a feasible one or to direct the particle to move towards feasible space. These methods usually start with initial feasible particles".

*Cut off at the boundary.* When a solution is moved to an infeasible location, its displacement vector is simply cut off. Either the solution is relocated on the boundary the nearest possible to the attempted new location, or the direction of the displacement vector is kept unmodified. Typically, the first alternative is preferred.

*Bisection method.* The cut off at the boundary technique works well when the boundaries are limited by interval constraints, and when the solution is located on the boundary. However, solutions might get trapped in the boundaries when the optimum is near them. Foryś et al.[19] proposed a *reflection technique* instead, while Innocente[15] proposed the *bisection method* (see Fig. 2): if the new position is infeasible, the displacement vector is split in halves, and the constraints verified. If the new position is still infeasible, the vector is split again, and so on. A feasible position will be found in time, unless the particle is already on the boundary. Hence an upper limit of iterations is set, and the particle keeps its position if no new feasible location can be found. Its drawbacks are that it cannot –in principle– pass through infeasible space, and that the particles must be initialized within $\mathcal{F}$. These drawbacks are shared to some extent with the *preserving feasibility* methods. Its main advantages are that it needs no adaptations to handle different inequality constraints, and it presents fast convergence (the latter being a double-edged sword).

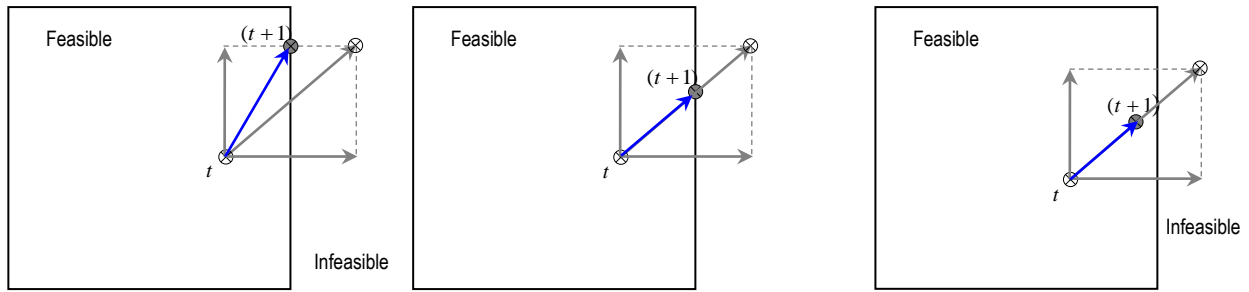

**Figure 2. Cut off at the boundary techniques (left and centre) and bisection method (right). On the left, the solution is relocated on the boundary the nearest possible to the attempted infeasible location. In the centre, the vector of displacement is cut off so that its direction remains unchanged. On the right, the vector of displacement is successively divided by two until the new solution is feasible.**

*4. Multiobjective-based techniques*

The main idea is to think of the constraints' violations as additional objectives to be minimized. Thus, constrained single-objective problems can be tackled using multiobjective optimization techniques.

The use of an evolutionary multiobjective optimization technique in this domain is not straightforward, because the number of objectives increases as we increase the number of constraints and there are not many evolutionary optimization techniques reported in the literature that have been accurately tested with more than a few objectives (normally no more than 5)[22]. Furthermore, the minimization of constraint violation while minimizing the value of the objective function is not as simple as it might seem. For example, if we concentrate first in just finding a feasible solution so that we can later concentrate in optimizing the objective function, then we would be sampling points in the feasible space at random and it would be then very difficult to approach the region when the optimum resides[23].

Numerous authors proposed solving the constrained single-objective problem using evolutionary algorithms equipped with multiobjective optimization techniques. Vieira et al.[32] reformulate constrained multiobjective problems by considering constraints as additional objectives and solving the resulting problem by the *Niched Pareto Genetic Algorithm* modified by incorporating the *Parks & Miller Elitist technique*. Note that the two test problems present only two objectives and one and two constraints, respectively. In turn, de Freitas Vaz et al.[31] propose an algo-rithm that uses a relaxed dominance concept adapted from multiobjective optimization. Thus, two objectives are sought: one is to minimize the objective function and the other is to obtain feasibility, the latter being more important[31]. A different approach, which does not use dominance to impose an order on the constraints based on their violation, is proposed by Coello Coello[21]. Instead, the objective function and the $m$ constraints are treated separately as $m+1$ objectives, each of which comprises the fitness function for a sub-population (i.e. the population is split into $m+1$ sub-populations). An ideal solution **x** would thus have $f_i(\mathbf{x})=0$ for $1 \leq i \leq m$ and $f(\mathbf{x}) \leq f(\mathbf{y})$ for all feasible **y** (assuming minimization)[21]. Other techniques consist of solving a bi-objective optimization problem, where individuals are Pareto-ranked based on the sum of constraints' violation. For an extensive review of multiobjective-based CHTs, refer to Mezura-Montes et al.[25, 27] and Coello Coello[24].

## IV. Features of the implemented algorithms

In order to compare different CHTs, the remaining features of the algorithm should remain the same. Nevertheless, it should be kept in mind that the same technique would perform different for different features of the algorithm. For instance, the bisection method quickly decreases the particles' momentum so that convergence occurs notably faster. Therefore this technique might work well with a local PSO, while premature convergence is likely with a global version of the algorithm. In order to allow replication of the experiments if desired, the features of the problem formulation and of the implemented algorithms are described hereafter.

### A. Formulation of the optimization problem

The optimization problem was formulated in Eqs. (1) and (2). It is self-evident that the PSO algorithm cannot comply with hard equality constraints unless some complex, problem-dependent repair algorithm is incorporated, or the problem is reformulated (e.g. G3 and G11 in Hu et al.[18]). However, such repair algorithm would be difficult to design, if possible at all. In turn, the reformulation of the problem to remove the equality constraints is also problem-dependent and not always possible (e.g. G5 in Hu et al.[18]). Therefore, equality constraints are relaxed here by setting a small tolerance, below which the solution is considered feasible despite the small constraints' violations. This is of common use in EAs to cope with these kinds of constraints. The implemented algorithm allows a different tolerance for equality and inequality constraints, where the interval constraints are viewed as inequality constraints. Thus, the optimization problem is reformulated as shown in Eq. (7):

$$\text{Minimize } f(\mathbf{x}), \text{ subject to } \begin{cases} g_j(\mathbf{x}) \leq Tolerance_{ineq} & ; \quad j=1,\ldots,q \\ \text{abs}(g_j(\mathbf{x})) \leq Tolerance_{eq} & ; \quad j=q+1,\ldots,m \\ \max(0, x_j - u_j) + \max(0, -x_j + l_j) \leq Tolerance_{ineq} & ; \quad j=1,\ldots,n \end{cases} \quad (7)$$

Thus, the amount of constraints' violation is given by Eq. (8):

$$cv = \sum_{j=1}^{q} \max(0, g_j(\mathbf{x})) + \sum_{j=q+1}^{m} \text{abs}(g_j(\mathbf{x})) + \sum_{j=1}^{n} \left[ \max(0, x_j - u_j) + \max(0, -x_j + l_j) \right] \quad (8)$$

### B. Features of the PSO algorithm

For the experiments carried out within this paper, we set $Tolerance_{ineq} = Tolerance_{eq} = 10^{-12}$. Note that most authors in the EAs literature typically set less demanding tolerances, between $10^{-3}$ and $10^{-5}$ (e.g. Muñoz Zavala et al.[34], de Freitas Vaz et al.[31], and Parsopoulos et al.[29], Takahama et al.[36, 37, 38], among others).

Kennedy et al.[7] suggest setting a population size between 10 and 50 particles, while Carlisle et al.[33] claim that a population size of 30 particles is a good choice. For the first test suite, we implemented 20 particles and 10000 time-steps so as to match the settings used by Hu et al.[17], whose results are used as a frame of reference for the first set of benchmark problems. For the second test suite, 30 particles and 8500 time-steps are implemented to match the settings in Toscano Pulido et al.[26], whose results are used as a frame of reference for this suite.

Nine CHTs were tested. Since different CHTs might result in different convergence rate, three neighbourhood sizes were considered, composed of 3, 11 and *swarm-size* particles, respectively. Note that the last neighbourhood size equals 20 for the first test suite and 40 for the second. The velocity constraint is set to half the dynamic range of the particles on each dimension. As to the coefficients of the velocity update equation, general-purpose settings were taken from previous studies[15]. Notice, however, that such settings were derived for the global version of the algo-rithm, aiming to balance its exploration and exploitation abilities. The settings consist of splitting the swarm in three, each of which is provided with different settings that result in complementary abilities:

*Sub-swarm 1*:    $w = 0.5$,    $iw = sw = 2$    (fine-cluster/fast convergence ability)
*Sub-swarm 2*:    $w = 0.7298$,    $iw = sw = 1.49609$    (fine-cluster/fast convergence ability)
*Sub-swarm 3*:    $w = 0.7$,    $iw = sw = 2$    (ability to escape poor solutions)

### C. Features of implemented CHTs

#### 1. *Preserving feasibility* (PF)

This is the same as originally proposed by Hu et al.[17], and its implementation is straightforward: the only modifications to the unconstrained PSO are that the random initialization of each particle is repeated until a feasible position is found, and that particles flying over infeasible space cannot store in memory the associated conflict function.

#### 2. *Preserving feasibility with priority rules* (PFPR)

A set of priority rules is incorporated to the original PF technique for the comparisons. Thus, priority is still given to the solution with the lower conflict if both solutions are feasible; and to the feasible solution if the other is infeasible. The innovation with respect to the original technique resides then in the comparison between two infeasible candidate solutions, where priority is given to the one with the lower constraints' violation (*cv*). The concept is straightforward and similar to the use of multiobjective techniques: in all constrained single-objective problems there are two objectives: the *minimization of the cost function* and the *maximization of the constraints' satisfaction*. The PFPR technique gives clear priority to the fulfilment of the second objective. This solves two of the major pitfalls of the original PF technique: an initial feasible swarm is no longer required, and some exploration of infeasible space is carried out. However, for problems with very low feasibility ratios –especially those with equality constraints–, most of the search is driven by constraints' satisfaction, completely disregarding the conflict function. This implies that by the time a particle finds a feasible location, it might be anywhere with respect to the optimal solution.

This technique is similar to that in Toscano Pulido et al.[26], although constraints' violations are calculated as in Eq. (8) without normalization. He et al.[30] also implemented a similar technique, called the *feasibility-based rule*.

#### 3. *Preserving feasibility with probabilistic priority rules* (PFPPR)

It basically consists of the PFPR but with the incorporation of a probability threshold in the priority rules when there is at least one infeasible solution involved in the comparison. That is to say, if both solutions are feasible, priority is still given to the one with the lower conflict value; if one solution is infeasible, there is a high probability (*prob*) that priority is given to the feasible particle, but there is a small chance (1–*prob*) that the comparison is based on their conflict values; finally, if both solutions are infeasible, there is a *prob* probability that priority is given to the one with the lower *cv*, while there is a (1–*prob*) chance that priority is given to the one with the lower conflict value. The aim is to improve the exploration of infeasible space, although it adds a new parameter to be tuned. We just set an arbitrary 90% probability threshold in our experiments, without further study.

It is noteworthy that the probability threshold is only used for the particles' best experience, whereas the plain priority rules are used in the comparisons to obtain the global and local best experiences.

#### 4. *Preserving feasibility with priority rules and relaxed equality constraints* (PFPR+REC)

Since we consider a very small tolerance for equality constraints' violations, the resulting feasibility ratio of the search space is always extremely small for problems with such constraints. Aiming to improve the exploration of infeasible space using the conflict function information rather than the constraints' violations, we proposed to significantly relax the tolerance at the first time-step, and gradually decrease it as the search progresses. While improvement is evident in the test problems with equality constraints, at least two new parameters to be tuned are added: the initial relaxed tolerance, and its rate of decrease.

Since we used the same number of time-steps for every problem in each test suite for comparison, we can easily make sure that the desired tolerance is reached before the search is terminated. Given that the objective is to increase the feasibility ratio, it makes sense to relate the initial relaxed tolerance to the space limited by the interval constraints. Nonetheless, the impact of a given relaxation on the feasibility ratio of the search space cannot, in principle, be predicted. The heuristic formulae used to compute the initial tolerance in our experiments is shown in Eq. (9). In order to give some time for the particles to find feasible solutions once the final tolerance ($10^{-12}$) is reached, it is arbitrarily set that such value is reached by the time 80% of the search has been carried out, as shown in Eq. (10).

$$Tolerance_{eq}^{ts=1} = \frac{\text{mean}(\mathbf{x}_{max} - \mathbf{x}_{min})}{2} \qquad (9)$$

$$Tolerance_{eq}^{ts \geq 0.8 t_{max}} = 10^{-12} \qquad (10)$$

Initially, we thought it made sense to decrease the tolerance exponentially, so that the rate of decrease is linked to the current value: $Tolerance_{eq}^{t} = 0.995 \cdot Tolerance_{eq}^{t-1}$. However, a linear decrease gave better results.

5. *Preserving feasibility with probabilistic priority rules and relaxed equality constraints* (PFPPR+REC)

It consists of the PFPPR technique, with the incorporation of the dynamic tolerance previously discussed.

6. *Penalization method* (PM)

We used Eqs. (5) and (6), where the penalization coefficients take the values shown in Eqs. (11) and (12). It is fair to note that this is not a much elaborated penalization mechanism, where not even the penalization coefficients have been studied and/or tuned. For an adaptive penalization scheme, refer to Parsopoulos et al.[29].

$$f_p(\mathbf{x}) = f(\mathbf{x}) + 10^6 \cdot \sum_{j=1}^{m} (f_j(\mathbf{x}))^2 \qquad (11)$$

$$f_j(\mathbf{x}) = \begin{cases} \max\{0, g_j(\mathbf{x})\} & ; \quad 1 \le j \le q \\ \text{abs}(g_j(\mathbf{x})) & ; \quad q < j \le m \end{cases} \qquad (12)$$

7. *Bisection method* (BM)

This is the same as originally proposed by Innocente[15]. The only modifications to the unconstrained PSO are that the random initialization of each particle is repeated until a feasible position is found, and that particles are not allowed to fly over infeasible space: if the new position is infeasible, the velocity vector is split in halves and feasibility is checked; if the relocation is still infeasible, the procedure is repeated until a feasible one is found.

8. *Bisection method with extra momentum* (BMEM)

Aiming to improve the BM, the decrease of particles' momentum is slowed down. If a particle attempts to move to an infeasible location, the velocity is stored in a temporary variable, and then adjusted by scaling it down by 0.9 rather than 0.5. If the new attempted location is still infeasible, the next position to try is given by scaling up the initial velocity (temporary variable) by 1.1. The process is repeated alternating scaling ups and downs until a feasible location is found. Note that only 19 trials are possible, after which the current location is maintained. This decreases the rate of decrease of the particles momentum, but initial feasible swarm is still required, and infeasible space is still not explored. For interval constraints, only scaling down is allowed.

9. *Bisection method with probabilistic extra momentum* (BMPEM)

Similar to the previous case, but the velocity adjustments are performed by multiplying the velocity by a random number $U_{(0,1.5)}$.

**D. Features of algorithms used by the authors considered in the comparisons**

1. *Hu et al.[17]*

Local PSO with neighbourhood size set to 3; $w = (0.5 + U_{(0,0.5)})$; $iw = sw = 1.49445$; 20 particles; 10000 time-steps yielding 200000 function evaluations (*fes*); velocity constraint set to the dynamic range of the particles on each dimension; and *Preserving Feasibility Technique* to handle constraints. 11 runs are performed for the statistics.

2. *Coello Coello[20]*

Co-evolutionary GA equipped with a *Self-Adaptive Penalization Method* to handle constraints. One population evolves solutions while another evolves the penalization coefficients. 11 runs are carried out for the statistics.

3. *He et al.[30]*

Hybrid PSO: global PSO + Simulated Annealing (SA); $w$ linearly decreasing from 0.9 to 0.4; $iw = sw = 2$; 250 particles; 300 time-steps; a *feasibility-based rule* to handle constraints (similar to our PFPR) and velocity constraint set to half the dynamic range of the particles on each dimension. A large population is set here to increase the parallel search, while the search does not need to be extended for fine-tuning as this is performed by a SA algorithm. Thus, a PSO is used to find a promising region while a SA to exploit it. 30 runs are performed for the statistics.

4. *Foryś et al.[19]*

A two-level sociality is implemented. Each particle is attracted towards a weighted average of its own best, its neighbourhood's best, and the swarm's best experiences. The neighbourhood architecture is updated at each time-step according to the nearest neighbours rule. The setting of the coefficients is not provided. They used a proposed reflexion technique for inequality and a penalization method with trapping mechanism for equality constraints.

5. *De Freitas Vaz et al.[31]*

Global PSO; 100 to 300 particles; $w$ linearly decreasing from 0.9 to 0.4; $iw = sw = 2$; a large number of *fes* different for each test problem; a tolerance of $10^{-4}$ for equality constraints; and a multiobjective-based CHT.

*6. Takahama et al.[38]*

ε Constrained PSO with Adaptive Velocity Limit Control (εPSO), which is a method that combines the ε Constrained Method with PSO; 20 particles and 2499 time-steps; $w$ linearly decreasing from 0.9 to 0.4; adaptive $v_{max}$ constraint; the constrained problem is turned into unconstrained by replacing ordinary comparisons by ε level comparisons; for the remaining parameters' settings, refer to Takahama et al.[38]. 30 runs are performed for the statistics.

*7. Toscano Pulido et al.[26]*

Global PSO with 40 particles; 8500 time-steps; $w = U_{(0.1,0.5)}$; $iw = U_{(1.5,2.5)}$; $sw = U_{(1.5,2.5)}$; we estimate they used a tolerance of $10^{-3}$ for equality constraints (inferred from the feasibility ratios in TABLE I[26], but not certain); turbulence operator incorporated to maintain diversity; and a mechanism similar to our PFPR is used to handle constraints (the only difference is that they *normalized* the constraints' violations). 30 runs are performed for the statistics.

*8. Hu et al.[18]*

Global PSO with 20 to 50 particles; 200 to 5000 time-steps (more for the second function in the second test suite); $w = (0.5 + U_{(0,0.5)})$; $iw = sw = 1.49445$; velocity constraint set to the dynamic range of the particles on each dimension; and PF technique to handle constraints. 20 runs are performed for the statistics.

*9. Parsopoulos et al.[29]*

Three variants of a global PSO are implemented: one with inertia weight, one with constriction factor, and one with both; $w$ linearly decreasing from 1.2 to 0.1; constriction factor $\chi = 0.73$; $iw = sw = 2$; 100 particles; 1000 time-steps; a dynamic penalization technique is used to handle constraints; $v_{max} = 4$; 10 runs performed for the statistics.

*10. Landa Becerra et al.[29]*

Cultured differential evolution algorithm consisting of a cultural algorithm coupled with a differential evolution algorithm as the search engine; same CHT as in Toscano Pulido et al.[26]; population of 100 individuals; 1000 generations; the settings of numerous parameters were derived empirically. 30 runs are performed for the statistics.

*11. Muñoz Zavala et al.[34]*

Particle Evolutionary Swarm Optimization (PESO), consisting of a global PSO with two perturbation operators to maintain diversity; 50 particles and 7000 time-steps yielding 350000 *fes*; no information given regarding the coefficients' settings; tolerance of $10^{-3}$ for equality constraints; a CHT similar to our PFPR (and to those in Toscano Pulido et al.[26] and in He et al.[30]). 30 runs carried out for the statistics.

*12. Zheng et al.[39]*

Improved Particle Swarm Optimization (IPSO), consisting of a multi-swarm PSO, where each sub-swarm is formed by proximity and has its own leader; a dynamic mutation operator is implemented to maintain diversity; 40 particles and 8500 time-steps yielding 340000 *fes*; $w = U_{(0.1,1)}$; $iw = U_{(1.5,2.5)}$; $sw = U_{(1.5,2.5)}$; a penalization method that discourage the exploration of infeasible space (the objective function is not involved in the calculation of the conflict of an infeasible particle) is implemented to handle constraints. 30 runs are performed for the statistics.

*13. Takahama et al.[36]*

α Constrained Genetic Algorithm (αGA), combining the α Constrained Method with a GA; 70 particles and 5000 time-steps; the constrained problem is turned into unconstrained by using the α level comparison; for the settings of other numerous parameters involved, refer to Takahama et al.[36]. 100 runs are performed for the statistics.

*14. Takahama et al.[37]*

Nonlinear Simplex Method with Mutations and α Constrained Method (αNSM); 90 search points and 85000 time-steps (8500 time-steps for 12$^{th}$ function in second test suite); the problem is turned into unconstrained using the α level comparison; a tolerance of $10^{-4}$ is used for equality constraints; refer to Takahama et al.[37] for the settings of other numerous parameters involved. 30 runs are performed for the statistics.

*15. Runarsson et al.[40]*

(30, 200)-ES with Stochastic Ranking selection and a penalization mechanism; tolerance of $10^{-4}$ for equality constraints; 200 offspring per generation, and 1750 generations yielding 350000 *fes*. 30 runs are performed for the statistics.

## V. Experimental results

Two suites of benchmark problems are used for the comparisons of the different CHTs implemented. Comparisons to other authors' results are provided as a frame of reference, although the main objective here is to compare the results obtained by the different CHTs and neighbourhood sizes, for which the remaining features of the algorithm should be the same. It is fair to remark, however, that the performance of a PSO algorithm depends on both the coefficients' settings and the CHT. Therefore it is possible to arrive to different conclusions for settings other than the general-purpose ones used here.

The feasibility ratios associated to each problem were calculated by randomly initializing $10^7$ particles within the interval constraints, and then calculating the percentage of them that were located on feasible space. Note that this was carried out with tolerances of $10^{-12}$ for both equality and inequality constraints in both test suites.

The first test suite is taken from Hu et al.[17], whose results are used as the main frame of reference. Hence the same number of particles, function evaluations (*fes*) and runs per sample (*nrs*) are kept: 20 particles, 10000 time-steps (200000 *fes*), and 11 runs. The second test suite is taken from Toscano Pulido et al.[26], whose results are used as the main frame of reference. Therefore the same number of particles, *fes* and runs per sample (*nrs*) are kept: 40 particles, 8500 time-steps (340000 *fes*), and 30 runs. For the formulation of these benchmark problems, refer to Hu et al.[17] and Toscano Pulido et al.[26]. They are not included here due to space constraints.

The best results in terms of both the best and mean conflicts found are highlighted to ease visualization in the tables. In turn, the very best results are further highlighted with darker background and white font. It is important to note that it is said here that a constraint is active when its tolerance is violated.

### A. First test suite of benchmark problems

It comprises five functions with inequality constraints. The first four functions are those in Hu et al.[17], whereas a continuous version of the Pressure Vessel Problem is added due to invalid comparisons performed by other authors.

*1. Pressure Vessel Design Problem*

The first problem in this test suite is the mixed-discrete Pressure Vessel Design Problem, which presents 4 dimensions, 3 inequality constraints, and a feasibility ratio of approximately 75%. The first two variables are integer multiples of 0.0625, and the other two variables are continuous. The discrete variables were handled by rounding-off the particles trajectories to the permitted discrete values. The experimental results are presented in Table 1.

| PROBLEM | | PRESSURE VESSEL (Mixed-discrete) | | | | | | | |
|---|---|---|---|---|---|---|---|---|---|
| | | Optimum | NI | NE | Dimensions | Feasibility ratio [%] | | | |
| | | - | 3 | 0 | 4 | 75.8937 | | | |
| | | BEST | | | MEAN | | | nrs | fes |
| CHT | nn | CONFLICT | nac | cv | CONFLICT | nac | cv | | |
| PF | 2 | 6059.714335 | 0 | 1.00E-12 | 6219.930401 | 0.00 | 2.71E-13 | 11 | 2.00E+05 |
| | 10 | 6090.526202 | 0 | 1.00E-12 | 6438.969348 | 0.00 | 1.00E-12 | 11 | 2.00E+05 |
| | 19 | 6090.526202 | 0 | 1.00E-12 | 6480.161969 | 0.00 | 9.09E-13 | 11 | 2.00E+05 |
| PFPR ≡ PFPR+REC | 2 | 6059.724449 | 0 | 0.00E+00 | 6234.257916 | 0.00 | 6.36E-13 | 11 | 2.00E+05 |
| | 10 | 6090.526202 | 0 | 1.00E-12 | 6357.933499 | 0.00 | 1.00E-12 | 11 | 2.00E+05 |
| | 19 | 6059.714335 | 0 | 1.00E-12 | 6314.746252 | 0.00 | 1.00E-12 | 11 | 2.00E+05 |
| PFPPR ≡ PFPPR+REC | 2 | 6090.526202 | 0 | 1.00E-12 | 6285.096682 | 0.00 | 1.00E-12 | 11 | 2.00E+05 |
| | 10 | 6370.779713 | 0 | 1.00E-12 | 6508.106316 | 0.00 | 1.00E-12 | 11 | 2.00E+05 |
| | 19 | 6090.526202 | 0 | 1.00E-12 | 6511.895813 | 0.00 | 1.00E-12 | 11 | 2.00E+05 |
| APM | 2 | 6039.708220 | 2 | 3.19E-03 | 6052.290983 | 1.36 | 2.71E-03 | 11 | 2.00E+05 |
| | 10 | 6073.242727 | 1 | 2.94E-03 | 6327.457958 | 1.73 | 2.80E-03 | 11 | 2.00E+05 |
| | 19 | 6040.229670 | 1 | 3.10E-03 | 6222.578595 | 1.73 | 2.72E-03 | 11 | 2.00E+05 |
| BM | 2 | 6059.714335 | 0 | 1.00E-12 | 6278.447090 | 0.00 | 1.00E-12 | 11 | 2.00E+05 |
| | 10 | 6059.714335 | 0 | 1.00E-12 | 6497.860196 | 0.00 | 1.00E-12 | 11 | 2.00E+05 |
| | 19 | 6059.714335 | 0 | 1.00E-12 | 6357.753566 | 0.00 | 1.00E-12 | 11 | 2.00E+05 |
| BMEM | 2 | 6059.714335 | 0 | 1.00E-12 | 6240.996653 | 0.00 | 1.00E-12 | 11 | 2.00E+05 |
| | 10 | 6059.714335 | 0 | 1.00E-12 | 6170.546711 | 0.00 | 1.00E-12 | 11 | 2.00E+05 |
| | 19 | 6059.714335 | 0 | 1.00E-12 | 6285.446198 | 0.00 | 1.00E-12 | 11 | 2.00E+05 |
| BMPEM | 2 | 6059.714335 | 0 | 1.00E-12 | 6310.797495 | 0.00 | 9.09E-13 | 11 | 2.00E+05 |
| | 10 | 6059.714335 | 0 | 1.00E-12 | 6635.585414 | 0.00 | 1.00E-12 | 11 | 2.00E+05 |
| | 19 | 6059.714335 | 0 | 1.00E-12 | 6553.468321 | 0.00 | 1.00E-12 | 11 | 2.00E+05 |
| Hu et al.[17] (local PSO) (*) | | 6059.131296 | - | - | - | - | - | 11 | 2.00E+05 |
| Coello Coello[20] (GA) | | 6288.744500 | - | - | 6293.848232 | - | - | 11 | - |
| He et al.[30] (PSO + SA) | | 6059.7143 | - | - | 6099.9323 | - | - | 30 | 8.10E+04 |
| Takahama et al.[38] (εPSO) | | 6059.7143 | - | - | 6154.4386 | - | - | 30 | 5.00E+04 |

(*) When replacing the reported object variables in their formulation, the conflict equals 6059.715172 rather than the 6059.131296 reported!

**Table 1. Best and mean conflicts found by our general-purpose PSO algorithm equipped with different CHTs, and those reported by other authors for reference, for the mixed-discrete Pressure Vessel Problem.**

The best conflict seems to converge towards 6059.714335. Although Hu et al.[17] reported a better result, the introduction of the corresponding variables reported into their formulation return a conflict value of 6059.715172. The penalization method also returns better results, but comparisons are unfair as it presents active constraints.

In terms of the best and the mean results obtained, it seems clear that the local versions of the algorithm perform better (notice that Hu et al.[17] chose a local PSO algorithm for this test suite, with 2 neighbours per particle). Although some constraints remain active, the best performance is exhibited by the additive penalization method with two neighbours per particle (i.e. 3-particle neighbourhood), whose mean and best conflicts are similar. This method still requires some refinement.

The priority rules, the probability threshold, the extra momentums did not result in clear improvement for this function, although it should be noticed that the feasibility ratio here is high.

Analyzing both best and mean conflicts, the results reported by He et al.[30] (PSO + SA) and Takahama et al.[38] (εPSO) are the best in Table 1.

Forys et al.[19] and de Freitas Vaz et al.[31] claimed to have found remarkably better results than those obtained by Hu et al.[17], but they did not seem to have considered the discrete nature of the first two variables. We added an extra function to this suite, consisting of the continuous pressure vessel design problem, and included their results as a frame of reference. The results are shown in Table 2.

| PROBLEM | | PRESSURE VESSEL (Continuous) | | | | | | | |
|---|---|---|---|---|---|---|---|---|---|
| | | Optimum | NI | NE | Dimensions | Feasibility ratio [%] | | | |
| | | - | 3 | 0 | 4 | 75.9314 | | | |
| | | BEST | | | MEAN | | | nrs | fes |
| CHT | nn | CONFLICT | nac | cv | CONFLICT | nac | cv | | |
| PF | 2 | 5887.357759 | 0 | 0.00E+00 | 5978.461133 | 0.00 | 1.33E-13 | 30 | 8.79E+05 |
| | 10 | 5894.090800 | 0 | 0.00E+00 | 5964.508424 | 0.00 | 6.67E-14 | 30 | 8.79E+05 |
| | 49 | 5906.564422 | 0 | 2.00E-12 | 6017.966362 | 0.00 | 1.39E-12 | 30 | 8.79E+05 |
| PFPR ≡ PFPR+REC | 2 | 5891.682631 | 0 | 0.00E+00 | 5955.424863 | 0.00 | 6.67E-14 | 30 | 8.79E+05 |
| | 10 | 5888.947679 | 0 | 0.00E+00 | 5929.624651 | 0.00 | 1.00E-13 | 30 | 8.79E+05 |
| | 49 | 5890.867453 | 0 | 2.73E-13 | 6013.753463 | 0.00 | 1.34E-12 | 30 | 8.79E+05 |
| PFPPR ≡ PFPPR+REC | 2 | 5895.582021 | 0 | 2.00E-12 | 6004.864562 | 0.00 | 2.00E-12 | 30 | 8.79E+05 |
| | 10 | 5912.976276 | 0 | 2.00E-12 | 6105.910337 | 0.00 | 2.00E-12 | 30 | 8.79E+05 |
| | 49 | 5896.273058 | 0 | 2.00E-12 | 6273.120851 | 0.00 | 2.00E-12 | 30 | 8.79E+05 |
| APM | 2 | 5849.942011 | 2 | 6.15E-03 | 5885.192852 | 2.33 | 4.42E-03 | 30 | 8.79E+05 |
| | 10 | 5854.796959 | 2 | 5.08E-03 | 5854.931854 | 3.83 | 5.07E-03 | 30 | 8.79E+05 |
| | 49 | 5854.933948 | 4 | 5.07E-03 | 5854.933953 | 3.97 | 5.07E-03 | 30 | 8.79E+05 |
| BM | 2 | 5886.761310 | 0 | 2.00E-12 | 5933.003826 | 0.00 | 1.95E-12 | 30 | 8.79E+05 |
| | 10 | 5886.126321 | 0 | 2.00E-12 | 5991.159384 | 0.00 | 1.97E-12 | 30 | 8.79E+05 |
| | 49 | 5886.776374 | 0 | 2.00E-12 | 6184.002832 | 0.00 | 2.00E-12 | 30 | 8.79E+05 |
| BMEM | 2 | 5885.795585 | 0 | 2.00E-12 | 5935.616137 | 0.00 | 2.00E-12 | 30 | 8.79E+05 |
| | 10 | 5886.122045 | 0 | 2.00E-12 | 5963.437962 | 0.00 | 2.00E-12 | 30 | 8.79E+05 |
| | 49 | 5887.394668 | 0 | 2.00E-12 | 6131.299871 | 0.00 | 2.00E-12 | 30 | 8.79E+05 |
| BMPEM | 2 | 5901.132304 | 0 | 0.00E+00 | 5967.216661 | 0.00 | 3.31E-14 | 30 | 8.79E+05 |
| | 10 | 5885.949297 | 0 | 0.00E+00 | 5998.710510 | 0.00 | 6.18E-14 | 30 | 8.79E+05 |
| | 49 | 5885.599799 | 0 | 2.00E-12 | 6094.285866 | 0.00 | 1.50E-12 | 30 | 8.79E+05 |
| de Freitas Vaz et al.[31] (global PSO + MO) | | 5885.33 | - | - | - | - | - | - | 8.79E+05 |
| Forys et al.[19] (local/global PSO) | | 5885.49 | - | - | - | - | - | - | - |

**Table 2. Best and mean conflicts found by our general-purpose PSO algorithm equipped with different CHTs, and those reported by other authors for reference, for the continuous Pressure Vessel Problem.**

This problem is actually better suited for the comparisons, as the PSO implemented here is for continuous functions. Thus, the rounding-off of the particles' trajectories is not affecting the performances. Notice that we used here the same *fes* as de Freitas Vaz et al.[31] for comparison, and a swarm of 30 particles.

Again, it can be observed that the local versions show better performances, and the APM find the best solutions in terms of both the best and the mean conflicts but cannot manage to find feasible solutions.

Here it can be seen that there is a slight improvement introduced by the priority rules, but not by the probability threshold. Also slight improvement seems to have result from the extra momentum (BMEM).

*2. Welded Beam Design Problem*

This is the second test problem in this suite, which presents 4 dimensions, 7 inequality constraints, and a feasibility ratio of approximately 2.65%. The experimental results are presented in Table 3, where the best results seem to converge to 1.724852. Again, Hu et al.[17] reported a result that seems to be slightly below the best solution, which may be due to some non-reported tolerance in their implementations, since their PF technique guarantees feasible solutions! It is fair to remark that the APM in Table 3 found the same result reported by Hu et al.[17] while exhibiting active constraints. Notice, however, that $cv < 10^{-6}$, which is an acceptable tolerance in many publications.

| PROBLEM | | WELDED BEAM DESIGN | | | | | | | |
|---|---|---|---|---|---|---|---|---|---|
| | | Optimum | NI | NE | Dimensions | Feasibility ratio [%] | | | |
| | | - | 7 | 0 | 4 | 2.6475 | | | |
| | | BEST | | | MEAN | | | nrs | fes |
| CHT | nn | CONFLICT | nac | cv | CONFLICT | nac | cv | | |
| PF | 2 | **1.724852** | **0** | **1.00E-12** | **1.724852** | **0.00** | 9.84E-13 | 11 | 2.00E+05 |
| | 10 | 1.724852 | 0 | 1.00E-12 | 1.725660 | 0.00 | 9.09E-13 | 11 | 2.00E+05 |
| | 19 | 1.724852 | 0 | 1.00E-12 | 1.724852 | 0.00 | 9.21E-13 | 11 | 2.00E+05 |
| PFPR ≡ PFPR+REC | 2 | **1.724852** | **0** | **1.00E-12** | **1.724852** | **0.00** | 1.00E-12 | 11 | 2.00E+05 |
| | 10 | 1.724852 | 0 | 1.00E-12 | 1.728946 | 0.00 | 9.09E-13 | 11 | 2.00E+05 |
| | 19 | 1.724852 | 0 | 1.00E-12 | 1.724878 | 0.00 | 7.63E-13 | 11 | 2.00E+05 |
| PFPPR ≡ PFPPR+REC | 2 | **1.724852** | **0** | **1.00E-12** | **1.724852** | **0.00** | 1.00E-12 | 11 | 2.00E+05 |
| | 10 | 1.724852 | 0 | 1.00E-12 | 1.756022 | 0.00 | 1.00E-12 | 11 | 2.00E+05 |
| | 19 | 1.724852 | 0 | 1.00E-12 | 1.930589 | 0.00 | 9.92E-13 | 11 | 2.00E+05 |
| APM | 2 | 1.724851 | 2 | 6.92E-07 | 1.724851 | 1.55 | 6.77E-07 | 11 | 2.00E+05 |
| | 10 | 1.724851 | 4 | 6.87E-07 | 1.724851 | 2.73 | 6.76E-07 | 11 | 2.00E+05 |
| | 19 | 1.724851 | 2 | 6.87E-07 | 1.724851 | 2.36 | 6.76E-07 | 11 | 2.00E+05 |
| BM | 2 | **1.724852** | **0** | **1.00E-12** | **1.724852** | **0.00** | 1.17E-12 | 11 | 2.00E+05 |
| | 10 | 1.724852 | 0 | 1.00E-12 | 1.725097 | 0.00 | 1.00E-12 | 11 | 2.00E+05 |
| | 19 | 1.724852 | 0 | 1.00E-12 | 1.774900 | 0.00 | 9.09E-13 | 11 | 2.00E+05 |
| BMEM | 2 | **1.724852** | **0** | **1.00E-12** | **1.724852** | **0.00** | 1.00E-12 | 11 | 2.00E+05 |
| | 10 | 1.724852 | 0 | 1.00E-12 | 1.725047 | 0.00 | 1.00E-12 | 11 | 2.00E+05 |
| | 19 | 1.724852 | 0 | 1.00E-12 | 1.725620 | 0.00 | 1.00E-12 | 11 | 2.00E+05 |
| BMPEM | 2 | **1.724852** | **0** | **1.00E-12** | 1.724898 | 0.00 | 9.92E-13 | 11 | 2.00E+05 |
| | 10 | 1.724852 | 0 | 1.00E-12 | 1.738441 | 0.00 | 1.00E-12 | 11 | 2.00E+05 |
| | 19 | 1.724852 | 0 | 1.91E-12 | 1.745821 | 0.00 | 1.08E-12 | 11 | 2.00E+05 |
| Hu et al.[17] (local PSO) (*) | | 1.724851 | - | - | - | - | - | 11 | 2.00E+05 |
| Coello Coello[20] (GA) | | 1.748309 | - | - | 1.771973 | - | - | 11 | - |
| de Freitas Vaz et al.[31] (global PSO + MO) | | 1.814290 | - | - | - | - | - | - | 9.60E+05 |
| Forys et al.[19] (local/global PSO) (#) | | 1.7248 | - | - | - | - | - | - | - |
| He et al.[30] (PSO + SA) | | **1.724852** | - | - | 1.749040 | - | - | 30 | 8.10E+04 |
| Takahama et al.[38] (εPSO) (#) | | 1.7249 | - | - | 1.7252 | - | - | 30 | 5.00E+04 |

(*) We suspect that the solution is below the optimum, but this is not definite.
(#) Not enough decimals for a fair comparison.

**Table 3. Best and mean conflicts found by our general-purpose PSO algorithm equipped with different CHTs, and those reported by other authors for reference, for the Welded Beam Design Problem.**

Although the best solution was found by all CHTs and neighbourhoods, the best performances are again exhibited by neighbourhoods composed of 3 particles (*nn* = 2). The APMs showed the best performances, but once again they did not manage to eliminate all constraints' violations.

The priority rules seem to be slightly harmful in terms of the results, although it must be kept in mind that for approximately the same results, the PFPR should always be preferred over the plain PF, given that the former does not require initial feasible swarm. In turn, the probability priority rules have not justified their existence yet!

As to the extra momentum, they improve the plain BM –especially for bigger neighbourhoods–, where the deterministic extra momentum (BMEM) gives better results than the probabilistic one (BMPEM).

*3. Tension/Compression Spring Design Problem*

This is the third test problem in Hu et al.'s benchmark suite[17], which presents 3 dimensions, 4 inequality constraints, and a feasibility ratio of approximately 0.75%. The experimental results are presented in Table 4, where the best results seem to converge to 0.12665.

Once again, $nn = 2$ leads to better results than bigger neighbourhoods. Although the APM still presents some constraints' violations, they are indeed small.

The PF with and without priority rules return similar results, which implies that the PFPR is preferable. The probabilistic threshold allow finding better *best conflicts*, although the mean conflicts are harmed. Overall, the PF and PFPR CHTs exhibit better performance.

| PROBLEM | | TENSION/COMPRESSION SPRING DESIGN | | | | | | | |
|---|---|---|---|---|---|---|---|---|---|
| | | Optimum | NI | NE | Dimensions | Feasibility ratio [%] | | | |
| | | - | 4 | 0 | 3 | 0.7467 | | | |
| | | BEST | | | MEAN | | | nrs | fes |
| CHT | nn | CONFLICT | nac | cv | CONFLICT | nac | cv | | |
| PF | 2 | 0.012671 | 0 | 0.00E+00 | 0.012767 | 0.00 | 3.62E-13 | 11 | 2.00E+05 |
| | 10 | 0.012685 | 0 | 0.00E+00 | 0.012883 | 0.00 | 7.98E-13 | 11 | 2.00E+05 |
| | 19 | 0.012668 | 0 | 2.00E-12 | 0.012861 | 0.00 | 8.83E-13 | 11 | 2.00E+05 |
| PFPR ≡ PFPR+REC | 2 | 0.012673 | 0 | 0.00E+00 | 0.012734 | 0.00 | 3.52E-13 | 11 | 2.00E+05 |
| | 10 | 0.012685 | 0 | 0.00E+00 | 0.012968 | 0.00 | 4.26E-13 | 11 | 2.00E+05 |
| | 19 | 0.012667 | 0 | 2.00E-12 | 0.012879 | 0.00 | 9.97E-13 | 11 | 2.00E+05 |
| PFPPR ≡ PFPPR+REC | 2 | 0.012667 | 0 | 2.00E-12 | 0.012975 | 0.00 | 2.00E-12 | 11 | 2.00E+05 |
| | 10 | 0.012666 | 0 | 2.00E-12 | 0.013195 | 0.00 | 2.00E-12 | 11 | 2.00E+05 |
| | 19 | 0.012681 | 0 | 2.00E-12 | 0.013528 | 0.00 | 2.00E-12 | 11 | 2.00E+05 |
| APM | 2 | 0.012668 | 0 | 0.00E+00 | 0.012796 | 0.55 | 2.11E-09 | 11 | 2.00E+05 |
| | 10 | 0.012689 | 2 | 1.73E-08 | 0.012935 | 1.18 | 8.27E-09 | 11 | 2.00E+05 |
| | 19 | 0.012679 | 2 | 1.76E-08 | 0.012888 | 1.45 | 1.19E-08 | 11 | 2.00E+05 |
| BM | 2 | 0.012665 | 0 | 1.00E-12 | 0.012750 | 0.00 | 1.71E-12 | 11 | 2.00E+05 |
| | 10 | 0.012666 | 0 | 2.00E-12 | 0.012812 | 0.00 | 2.00E-12 | 11 | 2.00E+05 |
| | 19 | 0.012667 | 0 | 2.00E-12 | 0.012940 | 0.00 | 2.00E-12 | 11 | 2.00E+05 |
| BMEM | 2 | 0.012665 | 0 | 0.00E+00 | 0.012812 | 0.00 | 1.27E-12 | 11 | 2.00E+05 |
| | 10 | 0.012665 | 0 | 2.00E-12 | 0.012902 | 0.00 | 2.00E-12 | 11 | 2.00E+05 |
| | 19 | 0.012676 | 0 | 2.00E-12 | 0.013132 | 0.00 | 2.00E-12 | 11 | 2.00E+05 |
| BMPEM | 2 | 0.012668 | 0 | 0.00E+00 | 0.012706 | 0.00 | 7.79E-15 | 11 | 2.00E+05 |
| | 10 | 0.012666 | 0 | 0.00E+00 | 0.012712 | 0.00 | 2.72E-13 | 11 | 2.00E+05 |
| | 19 | 0.012665 | 0 | 1.29E-12 | 0.012691 | 0.00 | 1.27E-12 | 11 | 2.00E+05 |
| Hu et al.[17] (local PSO) | | 0.012666 | - | - | - | - | - | 11 | 2.00E+05 |
| Coello Coello[20] (GA) | | 0.012705 | - | - | 0.012769 | - | - | 11 | - |
| de Freitas Vaz et al.[31] (global PSO + MO) | | 0.013193 | - | - | - | - | - | - | 7.58E+05 |
| He et al.[30] (PSO + SA) | | 0.012665 | - | - | 0.012707 | - | - | 30 | 8.10E+04 |

**Table 4. Best and mean conflicts found by our general-purpose PSO algorithm equipped with different CHTs, and those reported by other authors for reference, for the Tension/Compression Spring Design Problem.**

The bisection methods perform particularly well on this problem. The plain BM and the BMEM are at their best for $nn = 2$, whereas the BMPEM exhibit the best performance in the whole Table 4 (including other authors' results) for the global PSO ($nn = 19$). The reason for this is that the objective function itself is not a particularly hard one here (quadratic), and the difficulty of the problem is in the constraints. The bisection methods require initial feasible swarm, which means that at the initial time-step, all particles are already placed within the 0.75% of the search-space that is feasible, and none of them is allowed to ever leave such space. While this might be a drawback in other cases (e.g. harder objective function or disjointed feasible spaces), it certainly lead to good results for this problem.

4. Himmelblau's Nonlinear Problem

This is the last problem in Hu et al.'s benchmark suite[17], which presents 5 dimensions, 3 inequality constraints, and a feasibility ratio of approximately 52%. The experimental results are presented in Table 5, where the best results seem to converge to –31025.561420.

Not many conclusions can be derived from these results, as all CHTs find very good results. It can be observed, however, that $nn = 2$ returned the worst results for the PF and the PFPR techniques, whereas the global version did the same for the BM. This is because the BM itself favours fast convergence. Therefore convergence might just be too fast when combined with a global PSO. Besides, it appears that the extra momentums fixed that problem, and the global versions of the BMEM and BMPEM found the best solution. The APM again shows good performance in terms of consistency (best and mean solutions coincide), but it also confirms its inability to find feasible solutions.

| PROBLEM | | HIMMELBLAU'S NONLINEAR PROBLEM | | | | | | | |
|---|---|---|---|---|---|---|---|---|---|
| | | Optimum | NI | NE | Dimensions | Feasibility ratio [%] | | | |
| | | - | 3 | 0 | 5 | 52.0696 | | | |
| | | BEST | | | MEAN | | | nrs | fes |
| CHT | nn | CONFLICT | nac | cv | CONFLICT | nac | cv | | |
| PF | 2 | -31025.561416 | 0 | 0.00E+00 | -31025.557384 | 0.00 | 1.80E-13 | 11 | 2.00E+05 |
| | 10 | -31025.561420 | 0 | 4.86E-12 | -31025.561420 | 0.00 | 4.53E-12 | 11 | 2.00E+05 |
| | 19 | -31025.561420 | 0 | 4.91E-12 | -31025.561420 | 0.00 | 4.57E-12 | 11 | 2.00E+05 |
| PFPR ≡ PFPR+REC | 2 | -31025.561417 | 0 | 0.00E+00 | -31025.554504 | 0.00 | 2.90E-13 | 11 | 2.00E+05 |
| | 10 | -31025.561420 | 0 | 4.90E-12 | -31025.561420 | 0.00 | 4.68E-12 | 11 | 2.00E+05 |
| | 19 | -31025.561420 | 0 | 4.96E-12 | -31025.561420 | 0.00 | 4.49E-12 | 11 | 2.00E+05 |
| PFPPR ≡ PFPPR+REC | 2 | -31025.561420 | 0 | 4.81E-12 | -31025.561420 | 0.00 | 3.17E-12 | 11 | 2.00E+05 |
| | 10 | -31025.561420 | 0 | 4.84E-12 | -31025.561420 | 0.00 | 3.84E-12 | 11 | 2.00E+05 |
| | 19 | -31025.561420 | 0 | 4.33E-12 | -31025.561420 | 0.00 | 3.83E-12 | 11 | 2.00E+05 |
| APM | 2 | -31025.817789 | 5 | 4.99E-04 | -31025.817789 | 5.00 | 4.99E-04 | 11 | 2.00E+05 |
| | 10 | -31025.817789 | 5 | 4.99E-04 | -31025.817789 | 5.00 | 4.99E-04 | 11 | 2.00E+05 |
| | 19 | -31025.817789 | 5 | 4.99E-04 | -31025.817789 | 5.00 | 4.99E-04 | 11 | 2.00E+05 |
| BM | 2 | -31025.561420 | 0 | 1.99E-12 | -31025.561420 | 0.00 | 1.99E-12 | 11 | 2.00E+05 |
| | 10 | -31025.561420 | 0 | 1.99E-12 | -31025.561420 | 0.00 | 1.99E-12 | 11 | 2.00E+05 |
| | 19 | -31025.561420 | 0 | 1.99E-12 | -31025.561185 | 0.00 | 1.98E-12 | 11 | 2.00E+05 |
| BMEM | 2 | -31025.561420 | 0 | 1.99E-12 | -31025.561420 | 0.00 | 1.99E-12 | 11 | 2.00E+05 |
| | 10 | -31025.561420 | 0 | 1.99E-12 | -31025.561420 | 0.00 | 1.99E-12 | 11 | 2.00E+05 |
| | 19 | -31025.561420 | 0 | 1.99E-12 | -31025.561420 | 0.00 | 1.99E-12 | 11 | 2.00E+05 |
| BMPEM | 2 | -31025.561420 | 0 | 3.56E-12 | -31025.561420 | 0.00 | 2.48E-12 | 11 | 2.00E+05 |
| | 10 | -31025.561420 | 0 | 3.14E-12 | -31025.561420 | 0.00 | 2.20E-12 | 11 | 2.00E+05 |
| | 19 | -31025.561420 | 0 | 3.72E-12 | -31025.561420 | 0.00 | 2.56E-12 | 11 | 2.00E+05 |
| Hu et al.[17] (local PSO) | | -31025.561420 | - | - | - | - | - | 11 | 2.00E+05 |
| Coello Coello[20] (GA) (*) | | -31020.859000 | - | - | -30984.240703 | - | - | 11 | - |
| de Freitas Vaz et al.[31] (global PSO + MO) | | -31012.100000 | - | - | - | - | - | - | 7.84E+05 |
| Takahama et al.[38] (εPSO) | | -31025.5599 | - | - | -31025.5432 | - | - | 30 | 5.00E+04 |

(*) There is a difference in the 5th decimal of the term multiplying $x_1$ in the conflict function with respect to Hu et al.'s formulation[17].

**Table 5.** Best and mean conflicts found by our general-purpose PSO algorithm equipped with different CHTs, and those reported by other authors for reference, for the Himmelblau's Nonlinear Problem as formulated in Hu et al.[17].

With regards to comparisons with other authors' results, our optimizer finds the best result in terms of both the best and the mean conflict, with most of the CHTs and neighbourhood sizes tested. Hu et al.[17] also reported the same best result, although no statistics is provided.

**B. Second test suite of benchmark problems**

This suite was taken from Toscano Pulido et al.[26], and it comprises thirteen functions: nine with inequality constraints, three with equality constraints, and one function with both.

*1. Benchmark Problem 01*

This problem presents 13 dimensions, 9 inequality constraints, and a very small feasibility ratio of approximately 0.0003%. The experimental results are presented in Table 6, where the optimum equals –15.

All CHTs and neighbourhood sizes performed well regarding the best solution found, although the mean results obtained show that the local versions of the algorithm are more effective for this problem.

With regards to the CHT, it can be observed that the APM is unable to find solutions that do not violate some constraints beyond the tolerance. In turn, the priority rules seem to be slightly harmful in terms of the results (the plain PF find better results than the PFPR). However, it must be noticed that the random generation of a feasible initial swarm for this problem was extremely time-consuming, to the extent that it is concluded that the method is not suitable for problems with such small feasible ratios. The difficulty of this problem is in the small feasibility ratio, and not in the conflict function itself, which is only quadratic. With this in mind, it is not difficult to understand how a method that forces the whole swarm to be placed within the 0.0003% of the search-space that is feasible from the initial time step can find better solutions, especially when the drawback of the PFPR is that the exploration of infeasible space disregards the conflict value! Nevertheless, the local versions of PSO with PFPR perform reason-ably well, requiring remarkably less time than the plain PF. It is also not surprising to see that the probability threshold is beneficial in this problem, where maintain diversity for longer seems to be critical to find good results. Like the PF, the bisection methods require initial feasible swarm, and find very good results, especially for small neighbourhoods.

The results reported by other authors show that this problem does not pose much difficulty, and numerous optimizers in the literature are able to tackle it successfully.

| PROBLEM | | BENCHMARK PROBLEM 01 | | | | | | | |
|---|---|---|---|---|---|---|---|---|---|
| | | Optimum | NI | NE | Dimensions | Feasibility ratio [%] | | | |
| | | -15 | 9 | 0 | 13 | 0.0003 | | | |
| | | BEST | | | MEAN | | | nrs | fes |
| CHT | nn | CONFLICT | nac | cv | CONFLICT | nac | cv | | |
| PF | 2 | -14.992102 | 0 | 0.00E+00 | -14.975702 | 0.00 | 2.25E-14 | 30 | 3.40E+05 |
| | 10 | -14.998675 | 0 | 0.00E+00 | -14.986547 | 0.00 | 0.00E+00 | 30 | 3.40E+05 |
| | 39 | -15.000000 | 0 | 1.30E-11 | -14.585506 | 0.00 | 1.20E-11 | 30 | 3.40E+05 |
| PFPR ≡ PFPR+REC | 2 | -14.993046 | 0 | 0.00E+00 | -14.977214 | 0.00 | 1.33E-13 | 30 | 3.40E+05 |
| | 10 | -14.997890 | 0 | 0.00E+00 | -14.834239 | 0.00 | 6.52E-14 | 30 | 3.40E+05 |
| | 39 | -15.000000 | 0 | 1.20E-11 | -13.808579 | 0.00 | 1.11E-11 | 30 | 3.40E+05 |
| PFPPR ≡ PFPPR+REC | 2 | -15.000000 | 0 | 0.00E+00 | -15.000000 | 0.00 | 0.00E+00 | 30 | 3.40E+05 |
| | 10 | -15.000000 | 0 | 1.30E-11 | -14.933333 | 0.00 | 1.28E-11 | 30 | 3.40E+05 |
| | 39 | -15.000000 | 0 | 1.30E-11 | -13.750056 | 0.00 | 1.21E-11 | 30 | 3.40E+05 |
| APM | 2 | -14.995450 | 1 | 2.34E-05 | -14.981554 | 1.00 | 8.36E-06 | 30 | 3.40E+05 |
| | 10 | -15.000042 | 10 | 1.36E-05 | -14.930503 | 4.47 | 1.18E-05 | 30 | 3.40E+05 |
| | 39 | -15.000029 | 13 | 1.08E-05 | -14.387537 | 12.00 | 1.03E-05 | 30 | 3.40E+05 |
| BM | 2 | -15.000000 | 0 | 4.50E-12 | -15.000000 | 0.00 | 4.50E-12 | 30 | 3.40E+05 |
| | 10 | -15.000000 | 0 | 4.50E-12 | -14.960938 | 0.00 | 4.48E-12 | 30 | 3.40E+05 |
| | 39 | -15.000000 | 0 | 4.49E-12 | -14.177604 | 0.00 | 3.90E-12 | 30 | 3.40E+05 |
| BMEM | 2 | -15.000000 | 0 | 4.50E-12 | -15.000000 | 0.00 | 4.49E-12 | 30 | 3.40E+05 |
| | 10 | -15.000000 | 0 | 4.50E-12 | -14.788542 | 0.00 | 4.36E-12 | 30 | 3.40E+05 |
| | 39 | -15.000000 | 0 | 4.50E-12 | -14.218750 | 0.00 | 4.11E-12 | 30 | 3.40E+05 |
| BMPEM | 2 | -15.000000 | 0 | 7.99E-15 | -14.998758 | 0.00 | 1.18E-13 | 30 | 3.40E+05 |
| | 10 | -15.000000 | 0 | 9.47E-12 | -15.000000 | 0.00 | 6.41E-12 | 30 | 3.40E+05 |
| | 39 | -15.000000 | 0 | 6.33E-12 | -14.676468 | 0.00 | 4.32E-12 | 30 | 3.40E+05 |
| Toscano Pulido et al.[26] (global PSO) | | -15.000000 | - | - | -15.000000 | - | - | 30 | 3.40E+05 |
| Hu et al.[18] (global PSO + PF) (*) | | -15.0 | - | - | -15.0 | - | - | 20 | 2.50E+04 |
| Landa Becerra et al.[35] (CDE) | | -15.000000 | - | - | -14.999996 | - | - | 30 | 1.00E+05 |
| Muñoz Zavala et al.[34] (PESO) | | -15.000000 | - | - | -15.000000 | - | - | 30 | 3.50E+05 |
| Takahama et al.[36] (αGA) | | -15.000 | - | - | -15.000 | - | - | 100 | 3.50E+05 |
| Takahama et al.[37] (αNSM) | | -15.000000 | - | - | -14.999995 | - | - | 30 | 8.34E+04 |
| Runarsson et al.[40] (ES + SR) | | -15.000 | - | - | -15.000 | - | - | 30 | 3.50E+05 |

(*) Not enough decimals for a fair comparison.

**Table 6. Best and mean conflicts found by our general-purpose PSO algorithm equipped with different CHTs, and those reported by other authors for reference, for the Benchmark Problem 01 (g01) as formulated in Toscano Pulido et al.[26].**

*2. Benchmark Problem 02*

This problem presents 20 dimensions, 2 inequality constraints, and a very high feasibility ratio of approximately 99.99%. The experimental results are presented in Table 7, where the optimum equals –0.803619. The difficulty of this problem lies more in the conflict function than in the constraints.

With regards to the neighbourhood size, the best results are obtained for *nn* = 10 for all CHTs except for the Bisection methods. The latter show their best performance in this problem with a smaller neighbourhood (*nn* = 2). This is because the bisection methods quickly decrease the particles' momentum thus speeding up convergence, which is counterbalanced by a smaller neighbourhood (*nn* = 2).

Another phenomenon that can be observed is that the best solution is frequently better for a global algorithm in the preserving feasibility techniques at the expense of a poorer mean solution.

As it was expected, there is no difference between the PF with or without priority rules, since the feasibility ratio is very high. Therefore the search is driven by the conflict function value in both cases.

In a problem where the smaller neighbourhood work better with the BM because it maintains diversity for longer, it is reasonable to expect that the extra momentum would also beneficial as it is actually shown in Table 7.

| PROBLEM | | BENCHMARK PROBLEM 02 | | | | | | | |
|---|---|---|---|---|---|---|---|---|---|
| | | Optimum | NI | NE | Dimensions | Feasibility ratio [%] | | | |
| | | -0.803619 | 2 | 0 | 20 | 99.9964 | | | |
| | | BEST | | | MEAN | | | nrs | fes |
| CHT | nn | CONFLICT | nac | cv | CONFLICT | nac | cv | | |
| PF | 2 | -0.803030 | 0 | 9.99E-13 | -0.748428 | 0.00 | 3.00E-13 | 30 | 3.40E+05 |
| | 10 | **-0.803602** | **0** | **9.97E-13** | **-0.785558** | **0.00** | **3.30E-13** | 30 | 3.40E+05 |
| | 39 | -0.803617 | 0 | 7.80E-13 | -0.762096 | 0.00 | 3.26E-13 | 30 | 3.40E+05 |
| PFPR ≡ PFPR+REC | 2 | -0.803030 | 0 | 9.99E-13 | -0.748428 | 0.00 | 3.00E-13 | 30 | 3.40E+05 |
| | 10 | **-0.803602** | **0** | **9.97E-13** | **-0.785558** | **0.00** | **3.30E-13** | 30 | 3.40E+05 |
| | 39 | -0.803617 | 0 | 7.80E-13 | -0.762096 | 0.00 | 3.26E-13 | 30 | 3.40E+05 |
| PFPPR ≡ PFPPR+REC | 2 | -0.795679 | 0 | 0.00E+00 | -0.733225 | 0.00 | 0.00E+00 | 30 | 3.40E+05 |
| | 10 | -0.803613 | 0 | 0.00E+00 | -0.754572 | 0.00 | 0.00E+00 | 30 | 3.40E+05 |
| | 39 | -0.803607 | 0 | 4.84E-13 | -0.660676 | 0.00 | 2.00E-13 | 30 | 3.40E+05 |
| APM | 2 | -0.798329 | 0 | 0.00E+00 | -0.745892 | 0.33 | 7.20E-09 | 30 | 3.40E+05 |
| | 10 | **-0.803603** | **1** | **2.41E-08** | **-0.782393** | **0.67** | **1.21E-08** | 30 | 3.40E+05 |
| | 39 | -0.803616 | 0 | 0.00E+00 | -0.769005 | 0.60 | 1.10E-08 | 30 | 3.40E+05 |
| BM | 2 | -0.803603 | 0 | 9.96E-13 | -0.766311 | 0.00 | 6.70E-13 | 30 | 3.40E+05 |
| | 10 | -0.786122 | 0 | 1.00E-12 | -0.637235 | 0.00 | 9.99E-13 | 30 | 3.40E+05 |
| | 39 | -0.677159 | 0 | 1.00E-12 | -0.482020 | 0.00 | 9.98E-13 | 30 | 3.40E+05 |
| BMEM | 2 | **-0.803616** | **0** | **0.00E+00** | **-0.774361** | **0.00** | **9.99E-14** | 30 | 3.40E+05 |
| | 10 | -0.774021 | 0 | 1.00E-12 | -0.670290 | 0.00 | 9.99E-13 | 30 | 3.40E+05 |
| | 39 | -0.734076 | 0 | 9.99E-13 | -0.492969 | 0.00 | 9.74E-13 | 30 | 3.40E+05 |
| BMPEM | 2 | -0.800072 | 0 | 0.00E+00 | -0.775638 | 0.00 | 3.31E-14 | 30 | 3.40E+05 |
| | 10 | -0.799632 | 0 | 0.00E+00 | -0.734201 | 0.00 | 3.33E-14 | 30 | 3.40E+05 |
| | 39 | -0.786412 | 0 | 0.00E+00 | -0.619832 | 0.00 | 9.72E-14 | 30 | 3.40E+05 |
| Toscano Pulido et al.[26] (global PSO) | | **-0.803432** | - | - | **-0.790406** | - | - | 30 | 3.40E+05 |
| Hu et al.[18] (global PSO + PF) | | -0.8033 | - | - | -0.7521 | | | 20 | 5.00E+05 |
| Landa Becerra et al.[35] (CDE) | | **-0.803619** | - | - | -0.724886 | - | - | 30 | 1.00E+05 |
| Muñoz Zavala et al.[34] (PESO) | | -0.792608 | - | - | -0.721749 | - | - | 30 | 3.50E+05 |
| Zheng et al.[39] (IPSO) | | **-0.803491** | - | - | **-0.789935** | - | - | 30 | 3.40E+05 |
| Takahama et al.[37] (αNSM) | | **-0.803619** | - | - | **-0.784187** | - | - | 30 | 1.03E+05 |
| Runarsson et al.[40] (ES + SR) | | **-0.803515** | - | - | **-0.781975** | - | - | 30 | 3.50E+05 |

**Table 7. Best and mean conflicts found by our general-purpose PSO algorithm equipped with different CHTs, and those reported by other authors for reference, for the Benchmark Problem 02 (g02) as formulated in Toscano Pulido et al.[26].**

As to the results obtained by other authors in Table 7, those of Toscano Pulido et al.[26] (global PSO) and Takahama et al.[37] (αNSM) are undoubtedly the best in terms of both the best and mean conflicts.

*3. Benchmark Problem 03*

This problem presents 10 dimensions, 1 equality constraint, and a very small feasibility ratio (smaller than 0.0001%). The experimental results are presented in Table 8, where the optimum equals –1. The difficulty of this problem lies in both the conflict function than the equality constraint.

The equality constraint makes it virtually impossible to randomly generate an initial swarm, even in spite of the tolerance set for constraints' violations. As argued when discussing the results of Benchmark Problem 01, for feasibility ratios smaller than 0.0003%, we consider that a randomly generated initial feasible swarm is not possible. Hence the plain PF and the bisection methods are not valid alternatives when equality constraints are present. Since our objective is to develop a general CHT suitable for all problems, the plain PF is only used from here for as a frame of reference to make sure that the priority rules are not sensitively harmful. Similar, the bisection methods are still tested when possible to assess their performances for possible future improvements (aiming for some exploration of infeasible space).

The priority rules allow exploration of infeasible space, but disregarding the conflict function. For such small feasibility ratios, this implies that the particles are quickly pulled towards feasible space regardless of which part of it. Once a solution is feasible, the conflict function comes into the comparisons, but it may be too late to find promising areas within the feasible space. The results in Table 8 show that, although the PFPR and PFPPR are able to cope with equality constraints, the solutions found are not exactly impressive. It is a little bit surprising to see that good results were found by Toscano Pulido et al.[26], who used a global PSO algorithm with a CHT similar to our PFPR, although they proposed a normalization of the constraints' violations and different coefficients' settings (refer to section IV.D.*7*.). We did not study the reason for this yet, which could be the coefficients' settings, the normalization, or the turbulence operator. Nevertheless, given that we believe the reason for this poor performance is that the great majority of the search is driven by the constraints' violations rather than the conflict value, we proposed a dynamic relaxation of the tolerance (refer to section IV.C.*4*.).

| PROBLEM | | BENCHMARK PROBLEM 03 | | | | | | | |
|---|---|---|---|---|---|---|---|---|---|
| | | Optimum | NI | NE | Dimensions | Feasibility ratio [%] | | | |
| | | -1 | 0 | 1 | 10 | 0.0000 | | | |
| | | BEST | | | MEAN | | | nrs | fes |
| CHT | nn | CONFLICT | nac | cv | CONFLICT | nac | cv | | |
| PF | N/A | FAIL | | | | | | | |
| PFPR | 2 | -0.445397 | 0 | 1.00E-12 | -0.198607 | 0.00 | 9.28E-13 | 30 | 3.40E+05 |
| | 10 | -0.292403 | 0 | 1.00E-12 | -0.114537 | 0.00 | 8.76E-13 | 30 | 3.40E+05 |
| | 39 | -0.351423 | 0 | 9.13E-13 | -0.048849 | 0.00 | 9.05E-13 | 30 | 3.40E+05 |
| PFPPR | 2 | -0.503292 | 0 | 1.00E-12 | -0.191734 | 0.00 | 6.15E-13 | 30 | 3.40E+05 |
| | 10 | -0.428665 | 0 | 1.92E-13 | -0.129600 | 0.00 | 5.74E-13 | 30 | 3.40E+05 |
| | 39 | -0.252585 | 0 | 2.43E-13 | -0.061927 | 0.00 | 5.70E-13 | 30 | 3.40E+05 |
| PFPR+REC | 2 | **-0.994606** | **0** | **1.00E-12** | **-0.983277** | **0.00** | **9.70E-13** | 30 | 3.40E+05 |
| | 10 | **-0.997497** | **0** | **1.00E-12** | **-0.981756** | **0.00** | **9.85E-13** | 30 | 3.40E+05 |
| | 39 | -0.998461 | 0 | 1.00E-12 | -0.953667 | 0.00 | 3.38E-12 | 30 | 3.40E+05 |
| PFPPR+REC | 2 | -0.994101 | 0 | 1.00E-12 | -0.941992 | 0.00 | 8.91E-13 | 30 | 3.40E+05 |
| | 10 | -0.998317 | 0 | 9.98E-13 | -0.955979 | 0.00 | 7.70E-13 | 30 | 3.40E+05 |
| | 39 | -0.997837 | 0 | 9.93E-13 | -0.944807 | 0.00 | 6.61E-13 | 30 | 3.40E+05 |
| APM | 2 | -0.970041 | 1 | 1.37E-06 | -0.735194 | 1.00 | 4.17E-06 | 30 | 3.40E+05 |
| | 10 | **-0.997227** | 1 | 2.46E-06 | **-0.979747** | 1.00 | 8.95E-06 | 30 | 3.40E+05 |
| | 39 | **-1.000012** | 1 | 2.77E-06 | **-0.990626** | 1.00 | 3.26E-06 | 30 | 3.40E+05 |
| BM | N/A | FAIL | | | | | | | |
| BMEM | N/A | FAIL | | | | | | | |
| BMPEM | N/A | FAIL | | | | | | | |
| Toscano Pulido et al.[26] (global PSO) | | **-1.004720** | - | - | **-1.003814** | - | - | 30 | 3.40E+05 |
| Hu et al.[18] (global PSO + PF) (*) | | -1.0 | - | - | -1.0 | - | - | 20 | 1.00E+04 |
| Landa Becerra et al.[35] (CDE) | | -0.995413 | - | - | -0.788635 | - | - | 30 | 1.00E+05 |
| Muñoz Zavala et al.[34] (PESO) | | **-1.005010** | - | - | **-1.005006** | - | - | 30 | 3.50E+05 |
| Takahama et al.[37] (αNSM) | | **-1.000500** | - | - | **-1.000500** | - | - | 30 | 8.54E+04 |
| Runarsson et al.[40] (ES + SR) | | **-1.000** | - | - | **-1.000** | - | - | 30 | 3.50E+05 |

(*) The PF technique cannot handle equality constraints. A reformulation of the problem was carried out here to replace the equality constraint by an inequality one, so that the aim to test the algorithm against equality constraints is not performed!

**Table 8. Best and mean conflicts found by our general-purpose PSO algorithm equipped with different CHTs, and those reported by other authors for reference, for the Benchmark Problem 03 (g03) as formulated in Toscano Pulido et al.[26].**

This dynamic relaxation led to remarkable improvement, where the local versions and the PFPR+REC perform better than the global algorithm and the PFPPR+REC. In turn, the APM returned good results for the global PSO, although the constraint is still violated beyond the tolerance.

As to other authors' results, all of them found very good results except for Landa Becerra et al.[35] (CDE) and Hu et al.[18] (global PSO + PF). The latter did find the solution, but the problem was reformulated to remove the equality constraint, whereas the objective is not to show that the optimizer can solve a benchmark problem itself but that it can tackle the difficulties posed by it.

*4. Benchmark Problem 04*

This problem presents 5 dimensions, 3 inequality constraint, and a not very restrictive feasibility ratio of approximately 27%. The experimental results are presented in Table 9, where the optimum equals –30665.539.

| PROBLEM | | BENCHMARK PROBLEM 04 | | | | | | | |
|---|---|---|---|---|---|---|---|---|---|
| | | Optimum | NI | NE | Dimensions | Feasibility ratio [%] | | | |
| | | -30665.539 | 3 | 0 | 5 | 26.9552 | | | |
| | | BEST | | | MEAN | | | nrs | fes |
| CHT | nn | CONFLICT | nac | cv | CONFLICT | nac | cv | | |
| PF | 2 | -30665.538672 | 0 | 0.00E+00 | -30665.537801 | 0.00 | 3.01E-13 | 30 | 3.40E+05 |
| | 10 | -30665.538672 | 0 | 4.92E-12 | -30665.538672 | 0.00 | 4.88E-12 | 30 | 3.40E+05 |
| | 39 | -30665.538672 | 0 | 4.97E-12 | -30665.538672 | 0.00 | 4.87E-12 | 30 | 3.40E+05 |
| PFPR ≡ PFPR+REC | 2 | -30665.538669 | 0 | 0.00E+00 | -30665.537897 | 0.00 | 3.33E-13 | 30 | 3.40E+05 |
| | 10 | -30665.538672 | 0 | 4.91E-12 | -30665.538672 | 0.00 | 4.89E-12 | 30 | 3.40E+05 |
| | 39 | -30665.538672 | 0 | 4.97E-12 | -30665.538672 | 0.00 | 4.93E-12 | 30 | 3.40E+05 |
| PFPPR ≡ PFPPR+REC | 2 | -30665.538672 | 0 | 4.96E-12 | -30665.538672 | 0.00 | 3.63E-12 | 30 | 3.40E+05 |
| | 10 | -30665.538672 | 0 | 4.96E-12 | -30665.538672 | 0.00 | 4.90E-12 | 30 | 3.40E+05 |
| | 39 | -30665.538672 | 0 | 4.91E-12 | -30665.538672 | 0.00 | 4.69E-12 | 30 | 3.40E+05 |
| APM | 2 | -30665.952628 | 5 | 6.86E-04 | -30665.952627 | 5.00 | 6.86E-04 | 30 | 3.40E+05 |
| | 10 | -30665.952628 | 5 | 6.86E-04 | -30665.952627 | 5.00 | 6.86E-04 | 30 | 3.40E+05 |
| | 39 | -30665.952628 | 5 | 6.86E-04 | -30665.952627 | 5.00 | 6.86E-04 | 30 | 3.40E+05 |
| BM | 2 | -30665.538672 | 0 | 1.99E-12 | -30665.538672 | 0.00 | 1.99E-12 | 30 | 3.40E+05 |
| | 10 | -30665.538672 | 0 | 1.99E-12 | -30665.538672 | 0.00 | 1.99E-12 | 30 | 3.40E+05 |
| | 39 | -30665.538672 | 0 | 1.99E-12 | -30665.538672 | 0.00 | 1.99E-12 | 30 | 3.40E+05 |
| BMEM | 2 | -30665.538672 | 0 | 2.88E-12 | -30665.538672 | 0.00 | 2.14E-12 | 30 | 3.40E+05 |
| | 10 | -30665.538672 | 0 | 3.06E-12 | -30665.538672 | 0.00 | 2.08E-12 | 30 | 3.40E+05 |
| | 39 | -30665.538672 | 0 | 2.58E-12 | -30665.538672 | 0.00 | 2.20E-12 | 30 | 3.40E+05 |
| BMPEM | 2 | -30665.538672 | 0 | 2.78E-12 | -30665.526542 | 0.00 | 7.81E-13 | 30 | 3.40E+05 |
| | 10 | -30665.538672 | 0 | 4.74E-12 | -30665.538672 | 0.00 | 2.96E-12 | 30 | 3.40E+05 |
| | 39 | -30665.538672 | 0 | 3.93E-12 | -30665.538660 | 0.00 | 2.43E-12 | 30 | 3.40E+05 |
| Toscano Pulido et al.[26] (global PSO) | | -30665.500000 | - | - | -30665.500000 | - | - | 30 | 3.40E+05 |
| Hu et al.[18] (global PSO + PF) | | -30665.50 | - | - | -30665.50 | - | - | 20 | 1.00E+04 |
| He et al.[30] (PSO + SA) (#) | | -30665.539 | - | - | -30665.539 | - | - | 30 | 8.10E+04 |
| de Freitas Vaz et al.[31] (global PSO) | | -30665.50 | - | - | - | - | - | - | 9.75E+05 |
| Parsopoulos et al.[29] (global PSO+APM) | | -31544.459 | - | - | -31493.190 | - | - | 10 | 1.00E+05 |
| Landa Becerra et al.[35] (CDE) | | -30665.538672 | - | - | -30665.538672 | - | - | 30 | 1.00E+05 |
| Muñoz Zavala et al.[34] (PESO) | | -30665.538672 | - | - | -30665.538672 | - | - | 30 | 3.50E+05 |
| Takahama et al.[37] (αNSM) | | -30665.538672 | - | - | -30665.538672 | - | - | 30 | 7.41E+04 |
| Runarsson et al.[40] (ES + SR) (#) | | -30665.539 | - | - | -30665.539 | - | - | 30 | 3.50E+05 |

(#) Not enough decimals for a fair comparison.

**Table 9. Best and mean conflicts found by our general-purpose PSO algorithm equipped with different CHTs, and those reported by other authors for reference, for the Benchmark Problem 04 (g04) as formulated in Toscano Pulido et al.[26].**

Not many conclusions can be derived from these results, as all CHTs –as well as other authors' optimizers– exhibit good performance. It can be observed, however, that $nn = 2$ returned the worst results for the PF and the PFPR techniques, and that the priority rules have almost no effect in the results (therefore PFPR should be preferred over plain PF). Finally, the APM is again unable to eliminate all constraints' violations.

*5. Benchmark Problem 05*

This problem presents 4 dimensions, 1 inequality constraint, 3 equality constraints, and a very small feasibility ratio (smaller than 0.0001%). The experimental results are presented in Table 8, where the optimum equals 5126.498. The difficulty of this problem lies more in the constraints than in the conflict function.

Due to the extremely small feasibility ratio, the CHTs that require initial feasible swarm fail. As opposed to for the Benchmark Problem 03 (see Table 8), the PFPR and the PFPPR mange to find decent results despite the equality constraints (i.e. the very low feasibility ratio). This is believed to be due to the existence of more than one equality constraint (and lower dimensionality!), which narrow the feasible space even further. Thus, the difficulty is in finding a feasible solution rather than optimizing the feasible space. For that, the priority rules work well. Nevertheless, the best solution among the priority rules is that of the global PSO with PFPR+REC, followed by the PFPPR+REC. It can also be noticed that the local PSO algorithm seem to require more time to remove all infeasibilities. And as it could be expected, the APM found good results but still infeasible beyond the tolerance.

| PROBLEM | | BENCHMARK PROBLEM 05 | | | | | | | |
|---|---|---|---|---|---|---|---|---|---|
| | | Optimum | NI | NE | Dimensions | Feasibility ratio [%] | | | |
| | | 5126.498 | 1 | 3 | 4 | 0.0000 | | | |
| | | BEST | | | MEAN | | | nrs | fes |
| CHT | nn | CONFLICT | nac | cv | CONFLICT | nac | cv | | |
| PF | N/A | FAIL | | | | | | | |
| PFPR | 2 | 5126.981504 | 2 | 6.64E-03 | 5293.037892 | 2.50 | 1.43E-02 | 30 | 3.40E+05 |
| | 10 | 5127.788326 | 0 | 2.39E-12 | 5381.086741 | 1.43 | 1.49E-03 | 30 | 3.40E+05 |
| | 39 | 5126.805126 | 0 | 2.73E-12 | 5442.254518 | 0.30 | 6.49E-03 | 30 | 3.40E+05 |
| PFPPR | 2 | 5126.930692 | 0 | 2.39E-12 | 5380.300356 | 0.00 | 2.46E-12 | 30 | 3.40E+05 |
| | 10 | 5127.023157 | 0 | 2.50E-12 | 5277.439501 | 0.10 | 1.24E-03 | 30 | 3.40E+05 |
| | 39 | 5131.979757 | 0 | 2.50E-12 | 5436.587939 | 0.23 | 4.31E-03 | 30 | 3.40E+05 |
| PFPR+REC | 2 | 5126.477333 | 3 | 5.78E-03 | 5152.181415 | 2.83 | 1.10E-02 | 30 | 3.40E+05 |
| | 10 | 5127.586009 | 3 | 1.09E-08 | 5137.610955 | 2.20 | 1.45E-05 | 30 | 3.40E+05 |
| | 39 | 5126.500974 | 0 | 2.50E-12 | 5164.443416 | 0.40 | 2.77E-03 | 30 | 3.40E+05 |
| PFPPR+REC | 2 | 5126.527663 | 3 | 2.79E-08 | 5139.094881 | 2.83 | 1.11E-04 | 30 | 3.40E+05 |
| | 10 | 5128.349416 | 0 | 2.84E-12 | 5143.163497 | 0.03 | 6.66E-05 | 30 | 3.40E+05 |
| | 39 | 5126.534149 | 0 | 2.84E-12 | 5175.949509 | 0.47 | 5.57E-03 | 30 | 3.40E+05 |
| APM | 2 | 5126.491680 | 3 | 1.49E-02 | 5310.200941 | 3.10 | 2.15E-02 | 30 | 3.40E+05 |
| | 10 | 5127.221856 | 3 | 4.97E-04 | 5249.247655 | 3.23 | 4.43E-03 | 30 | 3.40E+05 |
| | 39 | 5126.732676 | 3 | 6.96E-06 | 5377.618057 | 3.20 | 1.30E-03 | 30 | 3.40E+05 |
| BM | N/A | FAIL | | | | | | | |
| BMEM | N/A | FAIL | | | | | | | |
| BMPEM | N/A | FAIL | | | | | | | |
| Toscano Pulido et al.[26] (global PSO) | | 5126.640000 | - | - | 5461.081333 | - | - | 30 | 3.40E+05 |
| Hu et al.[18] (global PSO + PF) | | FAIL | | | | | | | |
| Landa Becerra et al.[35] (CDE) | | 5126.570923 | - | - | 5207.410651 | - | - | 30 | 1.00E+05 |
| Muñoz Zavala et al.[34] (PESO) | | 5126.484154 | - | - | 5129.178298 | - | - | 30 | 3.50E+05 |
| Takahama et al.[37] ($\alpha$NSM) | | 5126.496714 | - | - | 5126.496714 | - | - | 30 | 6.50E+04 |
| Runarsson et al.[40] (ES + SR) | | 5126.497 | - | - | 5128.881 | - | - | 30 | 3.50E+05 |

**Table 10. Best and mean conflicts found by our general-purpose PSO algorithm equipped with different CHTs, and those reported by other authors for reference, for the Benchmark Problem 05 (g05) as formulated in Toscano Pulido et al.[26].**

As to the results obtained by other authors' implementations, the best performances are exhibited by those in Muñoz Zavala et al.[34] (PESO), Takahama et al.[37] ($\alpha$NSM), and Runarsson et al.[40] (ES + SR), although they appear to be slightly infeasible (according to the Optimum reported in Toscano Pulido et al.[26]).

*6. Benchmark Problem 06*

This problem presents 2 dimensions, 2 inequality constraints, and a fairly small feasibility ratio of approximately 0.007%. The experimental results are presented in Table 11, where the optimum equals –6961.81388.

Again, no useful conclusions can be drawn from this problem, most of the CHTs, neighbourhood sizes, and other authors' implementations were able to find good results. It can be seen, however, that the APM found itself again incapable of finding feasible solutions.

*7. Benchmark Problem 07*

This problem presents 10 dimensions, 8 inequality constraints, and a very small feasibility ratio of approximately 0.0001%. The experimental results are presented in Table 12, where the optimum equals 24.306209.

Due to the small feasibility ratio, the CHTs that require initial feasible swarm are out of the question despite the absence of equality constraints. The APM found decent results, but they are still infeasible. In contrast, the priority rules worked quite well, and reasonably good (feasible!) solutions were found with all three neighbourhood sizes, and with or without the probability threshold. Nonetheless, the best results are those corresponding to $nn = 10$.

With regards to the results found by other authors' optimizers, those of Landa Becerra et al.[35] (CDE) and of Takahama et al.[37] ($\alpha$NSM) exhibited the best performances overall.

| PROBLEM | | BENCHMARK PROBLEM 06 | | | | | | | |
|---|---|---|---|---|---|---|---|---|---|
| | | Optimum | NI | NE | Dimensions | Feasibility ratio [%] | | | |
| | | -6961.81388 | 2 | 0 | 2 | 0.0067 | | | |
| | | BEST | | | MEAN | | | nrs | fes |
| CHT | nn | CONFLICT | nac | cv | CONFLICT | nac | cv | | |
| PF | 2 | -6961.813876 | 0 | 8.24E-13 | -6961.813873 | 0.00 | 2.61E-13 | 30 | 3.40E+05 |
| | 10 | -6961.813876 | 0 | 1.99E-12 | -6961.813876 | 0.00 | 1.82E-12 | 30 | 3.40E+05 |
| | 39 | -6961.813876 | 0 | 1.99E-12 | -6961.813876 | 0.00 | 1.99E-12 | 30 | 3.40E+05 |
| PFPR ≡ PFPR+REC | 2 | -6961.813876 | 0 | 9.95E-13 | -6961.813873 | 0.00 | 2.63E-13 | 30 | 3.40E+05 |
| | 10 | -6961.813876 | 0 | 1.99E-12 | -6961.813876 | 0.00 | 1.70E-12 | 30 | 3.40E+05 |
| | 39 | -6961.813876 | 0 | 1.99E-12 | -6961.813876 | 0.00 | 1.99E-12 | 30 | 3.40E+05 |
| PFPPR ≡ PFPPR+REC | 2 | -6961.813876 | 0 | 1.99E-12 | -6961.813876 | 0.00 | 1.99E-12 | 30 | 3.40E+05 |
| | 10 | -6961.813876 | 0 | 1.99E-12 | -6961.813876 | 0.00 | 1.99E-12 | 30 | 3.40E+05 |
| | 39 | -6961.813876 | 0 | 1.99E-12 | -6961.813876 | 0.00 | 1.99E-12 | 30 | 3.40E+05 |
| APM | 2 | -6963.171126 | 2 | 1.16E-03 | -6963.171123 | 2.00 | 1.16E-03 | 30 | 3.40E+05 |
| | 10 | -6963.171126 | 2 | 1.16E-03 | -6963.171123 | 2.00 | 1.16E-03 | 30 | 3.40E+05 |
| | 39 | -6963.171128 | 2 | 1.16E-03 | -6963.171123 | 2.00 | 1.16E-03 | 30 | 3.40E+05 |
| BM | 2 | -6961.813876 | 0 | 1.99E-12 | -6961.813876 | 0.00 | 1.99E-12 | 30 | 3.40E+05 |
| | 10 | -6961.813876 | 0 | 1.99E-12 | -6961.813876 | 0.00 | 1.99E-12 | 30 | 3.40E+05 |
| | 39 | -6961.813876 | 0 | 1.99E-12 | -6961.813876 | 0.00 | 1.99E-12 | 30 | 3.40E+05 |
| BMEM | 2 | -6961.813876 | 0 | 1.99E-12 | -6961.813876 | 0.00 | 1.99E-12 | 30 | 3.40E+05 |
| | 10 | -6961.813876 | 0 | 1.99E-12 | -6961.813876 | 0.00 | 1.99E-12 | 30 | 3.40E+05 |
| | 39 | -6961.813876 | 0 | 1.99E-12 | -6961.813876 | 0.00 | 1.99E-12 | 30 | 3.40E+05 |
| BMPEM | 2 | -6961.813876 | 0 | 1.99E-12 | -6961.813876 | 0.00 | 1.99E-12 | 30 | 3.40E+05 |
| | 10 | -6961.813876 | 0 | 1.99E-12 | -6961.813876 | 0.00 | 1.99E-12 | 30 | 3.40E+05 |
| | 39 | -6961.813876 | 0 | 1.99E-12 | -6961.813876 | 0.00 | 1.99E-12 | 30 | 3.40E+05 |
| Toscano Pulido et al.[26] (global PSO) | | -6961.810000 | - | - | -6961.810000 | - | - | 30 | 3.40E+05 |
| Hu et al.[18] (global PSO + PF) (*) | | -6961.7 | - | - | -6960.7 | - | - | 20 | 1.00E+04 |
| de Freitas Vaz et al.[31] (global PSO) | | -6961.837 | - | - | - | - | - | - | 1.46E+06 |
| Parsopoulos et al.[29] (global PSO+APM) | | -6961.837 | - | - | -6961.774 | - | - | 10 | 1.00E+05 |
| Landa Becerra et al.[35] (CDE) | | -6961.813876 | - | - | -6961.813876 | - | - | 30 | 1.00E+05 |
| Muñoz Zavala et al.[34] (PESO) | | -6961.813876 | - | - | -6961.813876 | - | - | 30 | 3.50E+05 |
| Zheng et al.[39] (IPSO) | | -6961.814 | - | - | -6961.814 | - | - | 30 | 3.40E+05 |
| Takahama et al.[37] ($\alpha$NSM) | | -6961.813876 | - | - | -6961.813876 | - | - | 30 | 3.75E+04 |
| Runarsson et al.[40] (ES + SR) | | -6961.814 | - | - | -6875.940 | - | - | 30 | 3.50E+05 |

**Table 11.** Best and mean conflicts found by our general-purpose PSO algorithm equipped with different CHTs, and those reported by other authors for reference, for the Benchmark Problem 06 (g06) as formulated in Toscano Pulido et al.[26].

*8. Benchmark Problem 08*

This problem presents 2 dimensions, 2 inequality constraints, and a feasibility ratio of approximately 0.86%. The experimental results are presented in Table 13, where the optimum equals –0.095825. All CHTs with all neighbourhoods and all the optimizers implemented by other authors in Table 13, without exception, found the exact global optimum in every one of the runs carried out.

*9. Benchmark Problem 09*

This problem presents 7 dimensions, 4 inequality constraints, and a feasibility ratio of approximately 0.53%. The experimental results are presented in Table 14, where the optimum equals 680.630057.

Except for the BMPEM, all the other CHTs and neighbourhood sizes find reasonably good results. In particular, the preserving feasibility techniques find the best results, where the plain PF is only marginally superior to the PFPR, and the probability threshold is not beneficial. In turn, the APM find good results although some infeasibility is still present. The BM and BMEM find fairly good results, where the extra momentum appears to be marginally beneficial. Neighbourhood-wise, no advantage can be observed for such different neighbourhood sizes tested!

Regarding the best results in terms of both the best and mean solutions found reported by other authors, those in Landa Becerra et al.[35] (CDE), Muñoz Zavala et al.[34] (PESO), and Takahama et al.[37] (αNSM) are the best overall.

| PROBLEM | | BENCHMARK PROBLEM 07 | | | | | | | |
|---|---|---|---|---|---|---|---|---|---|
| | | Optimum | NI | NE | Dimensions | Feasibility ratio [%] | | | |
| | | 24.306209 | 8 | 0 | 10 | 0.0001 | | | |
| | | BEST | | | MEAN | | | nrs | fes |
| CHT | nn | CONFLICT | nac | cv | CONFLICT | nac | cv | | |
| PF | N/A | FAIL | | | | | | | |
| PFPR ≡ PFPR+REC | 2 | 24.450934 | 0 | 0.00E+00 | 24.719739 | 0.00 | 2.28E-13 | 30 | 3.40E+05 |
| | 10 | 24.323994 | 0 | 9.98E-13 | 24.735430 | 0.00 | 3.33E-14 | 30 | 3.40E+05 |
| | 39 | 24.336034 | 0 | 0.00E+00 | 25.012438 | 0.00 | 7.93E-13 | 30 | 3.40E+05 |
| PFPPR ≡ PFPPR+REC | 2 | 24.345401 | 0 | 0.00E+00 | 24.673363 | 0.00 | 0.00E+00 | 30 | 3.40E+05 |
| | 10 | 24.329901 | 0 | 4.93E-12 | 24.740877 | 0.00 | 4.56E-12 | 30 | 3.40E+05 |
| | 39 | 24.373750 | 0 | 4.96E-12 | 25.880024 | 0.00 | 4.66E-12 | 30 | 3.40E+05 |
| AP | 2 | 24.429805 | 1 | 8.95E-07 | 24.766104 | 0.43 | 5.85E-07 | 30 | 3.40E+05 |
| | 10 | 24.328603 | 1 | 1.48E-07 | 24.664268 | 0.87 | 1.09E-06 | 30 | 3.40E+05 |
| | 39 | 24.373697 | 5 | 2.19E-06 | 25.141183 | 3.53 | 2.08E-06 | 30 | 3.40E+05 |
| BM | N/A | FAIL | | | | | | | |
| BMEM | N/A | FAIL | | | | | | | |
| BMPEM | N/A | FAIL | | | | | | | |
| Toscano Pulido et al.[26] (global PSO) | | 24.341100 | - | - | 25.355771 | - | - | 30 | 3.40E+05 |
| Hu et al.[18] (global PSO + PF) | | 24.4420 | - | - | 26.7188 | - | - | 20 | 1.00E+05 |
| Landa Becerra et al.[35] (CDE) | | 24.306209 | - | - | 24.306210 | - | - | 30 | 1.00E+05 |
| Muñoz Zavala et al.[34] (PESO) | | 24.306921 | - | - | 24.371253 | - | - | 30 | 3.50E+05 |
| Takahama et al.[36] (αGA) | | 24.401 | - | - | 24.542 | - | - | 100 | 3.50E+05 |
| Takahama et al.[37] (αNSM) | | 24.306210 | - | - | 24.306263 | - | - | 30 | 8.61E+04 |
| Runarsson et al.[40] (ES + SR) | | 24.307 | - | - | 24.374 | - | - | 30 | 3.50E+05 |

**Table 12. Best and mean conflicts found by our general-purpose PSO algorithm equipped with different CHTs, and those reported by other authors for reference, for the Benchmark Problem 07 (g07) as formulated in Toscano Pulido et al.[26].**

*10. Benchmark Problem 10*

This problem presents 8 dimensions, 6 inequality constraints, and a very small feasibility ratio of approximately 0.0006%. The experimental results are presented in Table 15, where the optimum equals 7049.25. The conflict function is linear, but only in the first three out of eight variables.

The bisection methods do no perform very well on this problem, while the APM does in spite of not finding solutions with constraints' violations within the tolerance.

Here not only do the priority rules give the advantage of not requiring feasible initial swarm for such small feasibility ratio, but they also notably improve the solution. Given that all bisection methods and the plain PF technique perform the worst for this problem, it can be inferred that it is not convenient to use methods that require initial feasible swarm not only for the time-consuming initialization but also for the results found themselves! Both the PFPR and PFPPR exhibit reasonably good performances for their local versions, especially in terms of the mean solutions they were able to find.

With regards to the best results reported in the literature, those of Landa Becerra et al.[35] (CDE) and Takahama et al.[37] (αNSM) are indeed impressive.

*11. Benchmark Problem 11*

This problem presents 2 dimensions, 1 equality constraint, and a very small feasibility ratio of less than 0.0001%. The experimental results are presented in Table 16, where the optimum equals 0.75. The conflict function is only quadratic, so that the difficulty of this problem is obviously in handling the equality constraint. Therefore, the CHTs that require initial feasible swarm are out of the question.

The same as when optimizing the *Benchmark Problem 03*, the plain PFPR and PFPPR are able to find feasible solutions, as they perform some exploration of infeasible space. Although the best solutions found are reasonably accurate in this case, the mean solutions are still quite poor. The priority rules lead to remarkable improvement again, so that the PFPR+REC and the PFPPR+REC are able to find the best results in Table 16 in terms of both the

Preprint submitted to the 7th ASMO UK Conference on Engineering Design Optimizationbest and mean conflicts. The results do not seem to be sensitive to the neighbourhood size for these techniques when coping with this problem. The APM was able to find very good results, but the constraint remains active once the search has ended. The global neighbourhood appears more convenient for this method and this problem.

| PROBLEM | | BENCHMARK PROBLEM 08 | | | | | | | |
|---|---|---|---|---|---|---|---|---|---|
| | | Optimum | NI | NE | Dimensions | Feasibility ratio [%] | | | |
| | | -0.095825 | 2 | 0 | 2 | 0.8607 | | | |
| | | BEST | | | MEAN | | | nrs | fes |
| CHT | nn | CONFLICT | nac | cv | CONFLICT | nac | cv | | |
| PF | 2 | -0.095825 | 0 | 0.00E+00 | -0.095825 | 0.00 | 0.00E+00 | 30 | 3.40E+05 |
| | 10 | -0.095825 | 0 | 0.00E+00 | -0.095825 | 0.00 | 0.00E+00 | 30 | 3.40E+05 |
| | 39 | -0.095825 | 0 | 0.00E+00 | -0.095825 | 0.00 | 0.00E+00 | 30 | 3.40E+05 |
| PFPR ≡ PFPR+REC | 2 | -0.095825 | 0 | 0.00E+00 | -0.095825 | 0.00 | 0.00E+00 | 30 | 3.40E+05 |
| | 10 | -0.095825 | 0 | 0.00E+00 | -0.095825 | 0.00 | 0.00E+00 | 30 | 3.40E+05 |
| | 39 | -0.095825 | 0 | 0.00E+00 | -0.095825 | 0.00 | 0.00E+00 | 30 | 3.40E+05 |
| PFPPR ≡ PFPPR+REC | 2 | -0.095825 | 0 | 0.00E+00 | -0.095825 | 0.00 | 0.00E+00 | 30 | 3.40E+05 |
| | 10 | -0.095825 | 0 | 0.00E+00 | -0.095825 | 0.00 | 0.00E+00 | 30 | 3.40E+05 |
| | 39 | -0.095825 | 0 | 0.00E+00 | -0.095825 | 0.00 | 0.00E+00 | 30 | 3.40E+05 |
| APM | 2 | -0.095825 | 0 | 0.00E+00 | -0.095825 | 0.00 | 0.00E+00 | 30 | 3.40E+05 |
| | 10 | -0.095825 | 0 | 0.00E+00 | -0.095825 | 0.00 | 0.00E+00 | 30 | 3.40E+05 |
| | 39 | -0.095825 | 0 | 0.00E+00 | -0.095825 | 0.00 | 0.00E+00 | 30 | 3.40E+05 |
| BM | 2 | -0.095825 | 0 | 0.00E+00 | -0.095825 | 0.00 | 0.00E+00 | 30 | 3.40E+05 |
| | 10 | -0.095825 | 0 | 0.00E+00 | -0.095825 | 0.00 | 0.00E+00 | 30 | 3.40E+05 |
| | 39 | -0.095825 | 0 | 0.00E+00 | -0.095825 | 0.00 | 0.00E+00 | 30 | 3.40E+05 |
| BMEM | 2 | -0.095825 | 0 | 0.00E+00 | -0.095825 | 0.00 | 0.00E+00 | 30 | 3.40E+05 |
| | 10 | -0.095825 | 0 | 0.00E+00 | -0.095825 | 0.00 | 0.00E+00 | 30 | 3.40E+05 |
| | 39 | -0.095825 | 0 | 0.00E+00 | -0.095825 | 0.00 | 0.00E+00 | 30 | 3.40E+05 |
| BMPEM | 2 | -0.095825 | 0 | 0.00E+00 | -0.095825 | 0.00 | 0.00E+00 | 30 | 3.40E+05 |
| | 10 | -0.095825 | 0 | 0.00E+00 | -0.095825 | 0.00 | 0.00E+00 | 30 | 3.40E+05 |
| | 39 | -0.095825 | 0 | 0.00E+00 | -0.095825 | 0.00 | 0.00E+00 | 30 | 3.40E+05 |
| Toscano Pulido et al.[26] (global PSO) | | -0.0958 | - | - | -0.0958 | - | - | 30 | 3.40E+05 |
| Hu et al.[18] (global PSO + PF) | | -0.09583 | - | - | -0.09583 | - | - | 20 | 1.00E+04 |
| He et al.[30] (PSO + SA) | | -0.095825 | - | - | -0.095825 | - | - | 30 | 8.10E+04 |
| Landa Becerra et al.[35] (CDE) | | -0.095825 | - | - | -0.095825 | - | - | 30 | 1.00E+05 |
| Muñoz Zavala et al.[34] (PESO) | | -0.095825 | - | - | -0.095825 | - | - | 30 | 3.50E+05 |
| Zheng et al.[39] (IPSO) | | -0.095825 | - | - | -0.095825 | - | - | 30 | 3.40E+05 |
| Takahama et al.[37] (αNSM) | | -0.095825 | - | - | -0.095825 | - | - | 30 | 1.28E+05 |
| Runarsson et al.[40] (ES + SR) | | -0.095825 | - | - | -0.095825 | - | - | 30 | 3.50E+05 |

**Table 13.** Best and mean conflicts found by our general-purpose PSO algorithm equipped with different CHTs, and those reported by other authors for reference, for the Benchmark Problem 08 (g08) as formulated in Toscano Pulido et al.[26].

*12. Benchmark Problem 12*

This problem presents 3 dimensions, $9^3 = 729$ inequality constraints, and a feasibility ratio of approximately 4.8%. The experimental results are presented in Table 17, where the optimum equals –1. The conflict function is only quadratic, so that the difficulty of this problem is in handling disjointed feasible spaces, as the latter is composed of the aggregation of 729 disjointed spheres. Therefore it is arguable whether the problem presents 729 constraints or just one, as complying with only one of them implies feasibility while all the others are violated. However, all the constraints must not be satisfied for a particle to be infeasible.

With regards to the results obtained, all CHTs, all neighbourhood sizes, and all other authors' results reported and gathered in Table 17 are able to find the exact solution in every run. This means that the problem is not really challenging. Notice, for instance, that the bisection methods are, conceptually, awful in dealing with disjointed feasible spaces: particles are initialized within feasible space, and are not allowed to ever leave it. And yet, all of them found the exact solution, and for any neighbourhood size. This problem will be used for testing again in the future,

24
Association for Structural and Multidisciplinary Optimization in the UK (ASMO-UK)

but using much smaller diameters of the feasible spheres so that the feasibility ratio is much smaller. However, this will not allow comparisons to other authors' results.

| PROBLEM | | BENCHMARK PROBLEM 09 | | | | | | | |
|---|---|---|---|---|---|---|---|---|---|
| | | Optimum | NI | NE | Dimensions | Feasibility ratio [%] | | | |
| | | 680.630057 | 4 | 0 | 7 | 0.5264 | | | |
| | | BEST | | | MEAN | | | Nrs | fes |
| CHT | nn | CONFLICT | nac | cv | CONFLICT | nac | cv | | |
| PF | 2 | 680.633466 | 0 | 0.00E+00 | 680.642721 | 0.00 | 3.46E-13 | 30 | 3.40E+05 |
| | 10 | 680.631605 | 0 | 8.94E-13 | 680.637619 | 0.00 | 4.09E-13 | 30 | 3.40E+05 |
| | 39 | 680.630478 | 0 | 7.86E-13 | 680.640489 | 0.00 | 7.71E-13 | 30 | 3.40E+05 |
| PFPR ≡ PFPR+REC | 2 | 680.635232 | 0 | 0.00E+00 | 680.644817 | 0.00 | 2.16E-13 | 30 | 3.40E+05 |
| | 10 | 680.632118 | 0 | 9.47E-13 | 680.638264 | 0.00 | 3.15E-13 | 30 | 3.40E+05 |
| | 39 | 680.630794 | 0 | 9.99E-13 | 680.642411 | 0.00 | 6.94E-13 | 30 | 3.40E+05 |
| PFPPR ≡ PFPPR+REC | 2 | 680.634588 | 0 | 0.00E+00 | 680.670837 | 0.00 | 2.98E-13 | 30 | 3.40E+05 |
| | 10 | 680.636386 | 0 | 1.93E-12 | 680.677765 | 0.00 | 1.92E-12 | 30 | 3.40E+05 |
| | 39 | 680.660002 | 0 | 1.90E-12 | 680.770528 | 0.00 | 1.94E-12 | 30 | 3.40E+05 |
| APM | 2 | 680.635387 | 1 | 5.38E-07 | 680.649235 | 0.90 | 4.30E-07 | 30 | 3.40E+05 |
| | 10 | 680.632189 | 0 | 0.00E+00 | 680.638263 | 1.00 | 5.22E-07 | 30 | 3.40E+05 |
| | 39 | 680.631761 | 2 | 4.08E-07 | 680.648648 | 1.37 | 5.45E-07 | 30 | 3.40E+05 |
| BM | 2 | 680.630528 | 0 | 8.10E-13 | 680.679313 | 0.00 | 1.13E-12 | 30 | 3.40E+05 |
| | 10 | 680.633250 | 0 | 1.80E-12 | 680.680005 | 0.00 | 1.87E-12 | 30 | 3.40E+05 |
| | 39 | 680.634077 | 0 | 1.96E-12 | 680.716059 | 0.00 | 1.81E-12 | 30 | 3.40E+05 |
| BMEM | 2 | 680.631841 | 0 | 9.44E-13 | 680.656224 | 0.00 | 4.08E-13 | 30 | 3.40E+05 |
| | 10 | 680.634269 | 0 | 1.90E-12 | 680.655679 | 0.00 | 1.85E-12 | 30 | 3.40E+05 |
| | 39 | 680.631005 | 0 | 1.96E-12 | 680.672949 | 0.00 | 1.76E-12 | 30 | 3.40E+05 |
| BMPEM | 2 | 680.720626 | 0 | 0.00E+00 | 681.128281 | 0.00 | 0.00E+00 | 30 | 3.40E+05 |
| | 10 | 680.641893 | 0 | 0.00E+00 | 680.730116 | 0.00 | 3.68E-14 | 30 | 3.40E+05 |
| | 39 | 680.653774 | 0 | 0.00E+00 | 680.775075 | 0.00 | 1.05E-13 | 30 | 3.40E+05 |
| Toscano Pulido et al.[26] (global PSO) | | 680.638000 | - | - | 680.852393 | - | - | 30 | 3.40E+05 |
| Hu et al.[18] (global PSO + PF) | | 680.657 | - | - | 680.876 | - | - | 20 | 1.00E+04 |
| de Freitas Vaz et al.[31] (global PSO) | | 680.639000 | - | - | - | - | - | - | 1.69E+06 |
| Parsopoulos et al.[29] (global PSO+APM) | | 680.636 | - | - | 680.683 | - | - | 10 | 1.00E+05 |
| Landa Becerra et al.[35] (CDE) | | 680.630057 | - | - | 680.630057 | - | - | 30 | 1.00E+05 |
| Muñoz Zavala et al.[34] (PESO) | | 680.630057 | - | - | 680.630057 | - | - | 30 | 3.50E+05 |
| Takahama et al.[36] (αGA) | | 680.646 | - | - | 680.687 | - | - | 100 | 3.50E+05 |
| Takahama et al.[37] (αNSM) | | 680.630057 | - | - | 680.630057 | - | - | 30 | 8.51E+04 |
| Runarsson et al.[40] (ES + SR) | | 680.630 | - | - | 680.656 | - | - | 30 | 3.50E+05 |

**Table 14.** Best and mean conflicts found by our general-purpose PSO algorithm equipped with different CHTs, and those reported by other authors for reference, for the Benchmark Problem 09 (g09) as formulated in Toscano Pulido et al.[26].

*13. Benchmark Problem 13*

This is the last problem in the test suite taken from Toscano Pulido et al.[26]. It presents 5 dimensions, 3 equality constraints, and a feasibility ratio of less than 0.0001%. The experimental results are presented in Table 18, where the optimum equals 0.053950.

Due to the extremely small feasibility ratio, it is not possible to generate initial feasible swarm randomly. Hence the plain PF and bisection methods are off the table. While the priority rules –with or without probability threshold– make it possible to cope with such small feasibility ratio, the results are quite poor due to the fact that most of the search is carried out disregarding the conflict function. In fact, the local neighbourhoods with PFPR are not able to consistently find feasible solutions. The probability threshold gives some chance for infeasible particles to consider the conflict values during the search, and therefore the solutions found are improved. In addition, feasible solutions are found for every run and for every neighbourhood size. Nevertheless, the solutions are still poor.

Once again for problems with equality constraints, the dynamic relaxation of the tolerance is critical to find very good results. Thus, the PFPR+REC and the PFPPR+REC exhibit the best performance on this problem, which is one of the hardest among those included in the two test suites considered in this papers. In this regard, the local versions of the algorithm perform better in terms of the best and the mean solutions found, although they find it harder to comply with the tolerance permitted for the constraints' violations.

| PROBLEM | | BENCHMARK PROBLEM 10 | | | | | | | |
|---|---|---|---|---|---|---|---|---|---|
| | | Optimum | NI | NE | Dimensions | Feasibility ratio [%] | | | |
| | | 7049.25 | 6 | 0 | 8 | 0.0006 | | | |
| | | BEST | | | MEAN | | | nrs | fes |
| CHT | nn | CONFLICT | nac | cv | CONFLICT | nac | cv | | |
| PF | 2 | 7237.581424 | 0 | 0.00E+00 | 7483.440478 | 0.00 | 0.00E+00 | 30 | 3.40E+05 |
| | 10 | 7112.428290 | 0 | 0.00E+00 | 7405.230840 | 0.00 | 0.00E+00 | 30 | 3.40E+05 |
| | 39 | 7150.795503 | 0 | 3.00E-12 | 7419.915606 | 0.00 | 1.33E-12 | 30 | 3.40E+05 |
| PFPR ≡ PFPR+REC | 2 | 7090.641878 | 0 | 0.00E+00 | 7378.370307 | 0.00 | 0.00E+00 | 30 | 3.40E+05 |
| | 10 | 7057.952043 | 0 | 0.00E+00 | 7248.535972 | 0.00 | 3.29E-14 | 30 | 3.40E+05 |
| | 39 | 7064.199340 | 0 | 3.00E-12 | 7534.358675 | 0.00 | 2.05E-12 | 30 | 3.40E+05 |
| PFPPR ≡ PFPPR+REC | 2 | 7055.923237 | 0 | 0.00E+00 | 7246.513301 | 0.00 | 0.00E+00 | 30 | 3.40E+05 |
| | 10 | 7065.275001 | 0 | 3.00E-12 | 7376.626861 | 0.00 | 2.50E-12 | 30 | 3.40E+05 |
| | 39 | 7079.717916 | 0 | 3.00E-12 | 7699.279563 | 0.00 | 2.80E-12 | 30 | 3.40E+05 |
| APM | 2 | 7072.940031 | 3 | 1.17E-03 | 7254.589467 | 1.83 | 3.58E-03 | 30 | 3.40E+05 |
| | 10 | 7033.529336 | 3 | 5.46E-03 | 7224.850776 | 2.93 | 6.51E-03 | 30 | 3.40E+05 |
| | 39 | 7038.418965 | 4 | 5.99E-03 | 7330.216730 | 3.60 | 7.34E-03 | 30 | 3.40E+05 |
| BM | 2 | 7109.278278 | 0 | 7.40E+03 | 7395.779690 | 0.00 | 1.03E-12 | 30 | 3.40E+05 |
| | 10 | 7087.315710 | 0 | 2.75E-12 | 7348.226843 | 0.00 | 2.61E-12 | 30 | 3.40E+05 |
| | 39 | 7237.000025 | 0 | 2.00E-12 | 7737.340953 | 0.00 | 1.78E-12 | 30 | 3.40E+05 |
| BMEM | 2 | 7073.905025 | 0 | 0.00E+00 | 7302.386072 | 0.00 | 7.54E-13 | 30 | 3.40E+05 |
| | 10 | 7094.872918 | 0 | 3.00E-12 | 7342.867362 | 0.00 | 2.92E-12 | 30 | 3.40E+05 |
| | 39 | 7112.736008 | 0 | 2.00E-12 | 7527.782237 | 0.00 | 1.89E-12 | 30 | 3.40E+05 |
| BMPEM | 2 | 7132.496705 | 0 | 0.00E+00 | 7310.790475 | 0.00 | 0.00E+00 | 30 | 3.40E+05 |
| | 10 | 7122.326378 | 0 | 0.00E+00 | 7316.881880 | 0.00 | 4.37E-13 | 30 | 3.40E+05 |
| | 39 | 7112.047781 | 0 | 3.00E-12 | 7334.529736 | 0.00 | 2.66E-12 | 30 | 3.40E+05 |
| Toscano Pulido et al.[26] (global PSO) | | 7057.590000 | - | - | 7560.047857 | - | - | 30 | 3.40E+05 |
| Hu et al.[18] (global PSO + PF) | | 7131.010000 | - | - | 7594.650000 | - | - | 20 | 1.00E+05 |
| Landa Becerra et al.[35] (CDE) | | 7049.248058 | - | - | 7049.248266 | - | - | 30 | 1.00E+05 |
| Muñoz Zavala et al.[34] (PESO) | | 7049.459452 | - | - | 7099.101385 | - | - | 30 | 3.50E+05 |
| Takahama et al.[36] (αGA) | | 7053.951 | - | - | 7514.233 | - | - | 100 | 3.50E+05 |
| Takahama et al.[37] (αNSM) | | 7049.248021 | - | - | 7049.248022 | - | - | 30 | 8.01E+04 |
| Runarsson et al.[40] (ES + SR) | | 7054.316 | - | - | 7559.192 | - | - | 30 | 3.50E+05 |

**Table 15.** Best and mean conflicts found by our general-purpose PSO algorithm equipped with different CHTs, and those reported by other authors for reference, for the Benchmark Problem 10 (g10) as formulated in Toscano Pulido et al.[26].

With regards to the best solutions found and other results reported in the literature, it can be seen in Table 18 that the best results overall –considering our implementations only– is found by the local algorithm with $nn = 2$ and PFPPR+REC, which found a solution close to the optimum, and the best mean conflict in the whole of Table 18. However, some constraints' violations are still present in some runs (1.57 active constraints on average, although $cv < 1 \times 10^{-10}$). In turn, the global PSO with PFPPR+REC found the very best solution in the whole of Table 18, and no constraint remains active in any of the 30 runs. However, the mean conflict is not exactly outstanding. As to other authors' results, those reported in Takahama et al.[37] (αNSM) and Runarsson et al.[40] (ES + SR) are among the best ones, together with the PFPPR+REC with $nn = 2$.

## VI. Conclusion

The experimental results gathered in Table 1 to Table 18 show that different CHTs and neighbourhood sizes lead to better results for different problems, so that no optimum swarm-size or CHT could be extracted. Nonetheless, some conclusions could be drawn such as the inadmissibility of the initial feasible swarm requirement. Thus, the plain PF and the bisection methods are ruled out here in spite of their finding very good solutions in some problems, even when the feasibility ratio is very small. The latter is because those methods work well when the difficulty lies in the constraints rather than in the conflict function, so that the initial feasible swarm already implies having all particles placed within the extremely small feasible space before the search has even started. However, the random initialization of a feasible swarm becomes extremely time-consuming for highly constrained problems. The extra momentum incorporated to the bisection method resulted in alternated harmful and beneficial effects.

| PROBLEM | | BENCHMARK PROBLEM 11 | | | | | | | |
|---|---|---|---|---|---|---|---|---|---|
| | | Optimum | NI | NE | Dimensions | Feasibility ratio [%] | | | |
| | | 0.75 | 0 | 1 | 2 | 0.0000 | | | |
| | | BEST | | | MEAN | | | nrs | fes |
| CHT | nn | CONFLICT | nac | cv | CONFLICT | nac | cv | | |
| PF | N/A | FAIL | | | | | | | |
| PFPR | 2 | 0.750360 | 0 | 1.00E-12 | 0.762465 | 0.00 | 9.92E-13 | 30 | 3.40E+05 |
| | 10 | 0.751088 | 0 | 1.00E-12 | 0.849592 | 0.00 | 9.71E-13 | 30 | 3.40E+05 |
| | 39 | 0.754376 | 0 | 1.00E-12 | 0.867122 | 0.00 | 8.71E-13 | 30 | 3.40E+05 |
| PFPPR | 2 | 0.750005 | 0 | 1.00E-12 | 0.783429 | 0.00 | 8.81E-13 | 30 | 3.40E+05 |
| | 10 | 0.755515 | 0 | 9.28E-13 | 0.918178 | 0.00 | 6.63E-13 | 30 | 3.40E+05 |
| | 39 | 0.750315 | 0 | 1.00E-12 | 0.844219 | 0.00 | 5.88E-13 | 30 | 3.40E+05 |
| PFPR+REC | 2 | 0.750000 | 0 | 1.00E-12 | 0.750049 | 0.00 | 9.51E-13 | 30 | 3.40E+05 |
| | 10 | 0.750000 | 0 | 1.00E-12 | 0.750033 | 0.00 | 9.96E-13 | 30 | 3.40E+05 |
| | 39 | 0.750000 | 0 | 1.00E-12 | 0.750210 | 0.00 | 9.97E-13 | 30 | 3.40E+05 |
| PFPPR+REC | 2 | 0.750000 | 0 | 1.00E-12 | 0.750089 | 0.00 | 1.00E-12 | 30 | 3.40E+05 |
| | 10 | 0.750000 | 0 | 1.00E-12 | 0.750061 | 0.00 | 9.93E-13 | 30 | 3.40E+05 |
| | 39 | 0.750000 | 0 | 1.00E-12 | 0.750076 | 0.00 | 9.96E-13 | 30 | 3.40E+05 |
| APM | 2 | 0.749997 | 1 | 2.91E-06 | 0.750899 | 1.00 | 6.13E-07 | 30 | 3.40E+05 |
| | 10 | 0.749999 | 1 | 5.00E-07 | 0.750015 | 1.00 | 5.01E-07 | 30 | 3.40E+05 |
| | 39 | 0.750000 | 1 | 5.00E-07 | 0.750000 | 1.00 | 5.07E-07 | 30 | 3.40E+05 |
| BM | N/A | FAIL | | | | | | | |
| BMEM | N/A | FAIL | | | | | | | |
| BMPEM | N/A | FAIL | | | | | | | |
| Toscano Pulido et al.[26] (global PSO) | | 0.749999 | - | - | 0.750107 | - | - | 30 | 3.40E+05 |
| Hu et al.[18] (global PSO + PF) (*) | | 0.75 | - | - | 0.75 | - | - | 20 | 1.00E+04 |
| Landa Becerra et al.[35] (CDE) | | 0.749900 | - | - | 0.757995 | - | - | 30 | 1.00E+05 |
| Muñoz Zavala et al.[34] (PESO) | | 0.749000 | - | - | 0.749000 | - | - | 30 | 3.50E+05 |
| Zheng et al.[39] (IPSO) | | 0.750 | - | - | 0.750 | - | - | 30 | 3.40E+05 |
| Takahama et al.[37] (αNSM) | | 0.749900 | - | - | 0.749900 | - | - | 30 | 8.49E+04 |
| Runarsson et al.[40] (ES + SR) | | 0.750 | - | - | 0.750 | - | - | 30 | 3.50E+05 |

(*) The PF technique cannot handle equality constraints. A reformulation of the problem was carried out here to replace the equality constraint by an inequality one, so that the aim to test the algorithm against equality constraints is not performed!

**Table 16. Best and mean conflicts found by our general-purpose PSO algorithm equipped with different CHTs, and those reported by other authors for reference, for the Benchmark Problem 11 (g11) as formulated in Toscano Pulido et al.[26].**

The penalization method used in our experiments led to good results in general but showed itself incapable of finding feasible solutions, even when some tolerance for the constraints' violation was accepted. That is to say, most of its solutions presented constraints' violations beyond the tolerance. It must be noted, however, that the penalization coefficients were set without tuning, and the penalization method itself was rather basic. Some improvements are required such as adaptive penalization, or at least setting $\alpha_j = 1$ when $f_j(\mathbf{x}) < 1$.

The preserving feasibility technique with priority rules proved itself quite robust, but it presented some problems for very low feasibility ratios –especially when equality constraints were involved– due to the fact that most of the search was driven by the constraints' violations, disregarding the conflict function. This problem is more important in problems where the difficulty lies in both the constraints and the conflict function. The probability threshold introduced aimed at improving the exploration of infeasible space by considering the conflict value of infeasible solutions with some probability. However, improvement was not consistent. Therefore we incorporated a dynamic (time-decreasing) relaxation of the tolerance for equality constraints' violations, so that the search was driven by the conflict function at the early stages even when the PFPR technique was used. This resulted in remarkable improvement, although the performance was sensitive to the initial relaxation and to the rate of decrease. We tried linear and exponential decrease, where the former led to better results.

| PROBLEM | | BENCHMARK PROBLEM 12 | | | | | | | |
|---|---|---|---|---|---|---|---|---|---|
| | | Optimum | NI | NE | Dimensions | Feasibility ratio [%] | | | |
| | | -1 | 729 (*) | 0 | 3 | 4.7713 | | | |
| | | BEST | | | MEAN | | | | |
| CHT | nn | CONFLICT | nac | cv | CONFLICT | nac | cv | nrs | fes |
| PF | 2 | -1.000000 | 0 | 0.00E+00 | -1.000000 | 0.00 | 0.00E+00 | 30 | 3.40E+05 |
| | 10 | -1.000000 | 0 | 0.00E+00 | -1.000000 | 0.00 | 0.00E+00 | 30 | 3.40E+05 |
| | 39 | -1.000000 | 0 | 0.00E+00 | -1.000000 | 0.00 | 0.00E+00 | 30 | 3.40E+05 |
| PFPR ≡ PFPR+REC | 2 | -1.000000 | 0 | 0.00E+00 | -1.000000 | 0.00 | 0.00E+00 | 30 | 3.40E+05 |
| | 10 | -1.000000 | 0 | 0.00E+00 | -1.000000 | 0.00 | 0.00E+00 | 30 | 3.40E+05 |
| | 39 | -1.000000 | 0 | 0.00E+00 | -1.000000 | 0.00 | 0.00E+00 | 30 | 3.40E+05 |
| PFPPR ≡ PFPPR+REC | 2 | -1.000000 | 0 | 0.00E+00 | -1.000000 | 0.00 | 0.00E+00 | 30 | 3.40E+05 |
| | 10 | -1.000000 | 0 | 0.00E+00 | -1.000000 | 0.00 | 0.00E+00 | 30 | 3.40E+05 |
| | 39 | -1.000000 | 0 | 0.00E+00 | -1.000000 | 0.00 | 0.00E+00 | 30 | 3.40E+05 |
| APM | 2 | -1.000000 | 0 | 0.00E+00 | -1.000000 | 0.00 | 0.00E+00 | 30 | 3.40E+05 |
| | 10 | -1.000000 | 0 | 0.00E+00 | -1.000000 | 0.00 | 0.00E+00 | 30 | 3.40E+05 |
| | 39 | -1.000000 | 0 | 0.00E+00 | -1.000000 | 0.00 | 0.00E+00 | 30 | 3.40E+05 |
| BM | 2 | -1.000000 | 0 | 0.00E+00 | -1.000000 | 0.00 | 0.00E+00 | 30 | 3.40E+05 |
| | 10 | -1.000000 | 0 | 0.00E+00 | -1.000000 | 0.00 | 0.00E+00 | 30 | 3.40E+05 |
| | 39 | -1.000000 | 0 | 0.00E+00 | -1.000000 | 0.00 | 0.00E+00 | 30 | 3.40E+05 |
| BMEM | 2 | -1.000000 | 0 | 0.00E+00 | -1.000000 | 0.00 | 0.00E+00 | 30 | 3.40E+05 |
| | 10 | -1.000000 | 0 | 0.00E+00 | -1.000000 | 0.00 | 0.00E+00 | 30 | 3.40E+05 |
| | 39 | -1.000000 | 0 | 0.00E+00 | -1.000000 | 0.00 | 0.00E+00 | 30 | 3.40E+05 |
| BMPEM | 2 | -1.000000 | 0 | 0.00E+00 | -1.000000 | 0.00 | 0.00E+00 | 30 | 3.40E+05 |
| | 10 | -1.000000 | 0 | 0.00E+00 | -1.000000 | 0.00 | 0.00E+00 | 30 | 3.40E+05 |
| | 39 | -1.000000 | 0 | 0.00E+00 | -1.000000 | 0.00 | 0.00E+00 | 30 | 3.40E+05 |
| Toscano Pulido et al.[26] (global PSO) | | -1.000000 | - | - | -1.000000 | - | - | 30 | 3.40E+05 |
| Hu et al.[18] (global PSO + PF) | | -1.0 | - | - | -1.0 | - | - | 20 | 1.00E+04 |
| He et al.[30] (PSO + SA) | | -1.000000 | - | - | -1.000000 | - | - | 30 | 8.10E+04 |
| Landa Becerra et al.[35] (CDE) | | -1.000000 | - | - | -1.000000 | - | - | 30 | 1.00E+05 |
| Muñoz Zavala et al.[34] (PESO) | | -1.000000 | - | - | -1.000000 | - | - | 30 | 3.50E+05 |
| Takahama et al.[37] (αNSM) | | -1.000000 | - | - | -1.000000 | - | - | 30 | 1.35E+04 |
| Runarsson et al.[40] (ES + SR) | | -1.000 | - | - | -1.000 | - | - | 30 | 3.50E+05 |

(*) The number of inequality constraints is arguable. It is typically claimed that there are $9^3$ inequality constraints because the feasible space consists of that number of disjointed spheres, and all of them must be checked to confirm infeasibility. However, only one of them has to (and can) be true for a particle to be feasible, so that it can be said that all the spheres in reality comprise a single stepwise constraint.

**Table 17. Best and mean conflicts found by our general-purpose PSO algorithm equipped with different CHTs, and those reported by other authors for reference, for the Benchmark Problem 12 (g12) as formulated in Toscano Pulido et al.[26].**

*Future work*

The coming work for the immediate future involves the development of dynamic neighbourhoods, improvement of some of the constraint-handling techniques tested, development and test of diversity operators, and a systematic study of the main parameters of the velocity update equations (extending the one carried out by Innocente[15]).

One of the first important conclusions that can be derived from the optimization of all benchmark functions in both test suites is that the most convenient neighbourhood size is problem-dependent. A global neighbourhood speeds up convergence, and helps ensuring that feasible solutions are found and that the search is fine-tuned. In contrast, more local versions help avoiding premature convergence, but the latter might end up being so slow in some problems that fine-tuning of the search –and fine-clustering of the particles– does not take place. Thus, compliance with constraints' is more likely to fail. Therefore a dynamic neighbourhood size will be implemented, where the neighbourhood-size increases as the search progresses. Other issues will be in time studied such as the effect of choosing the neighbourhood according to the nearest neighbour rule both at the initial time-step only and on every time-step; adaptive neighbourhoods related to the convergence; and adaptive neighbourhoods according to the actual distance in the search space (defining areas of influence).

| PROBLEM | | BENCHMARK PROBLEM 13 | | | | | | | |
|---|---|---|---|---|---|---|---|---|---|
| | | Optimum | NI | NE | Dimensions | Feasibility ratio [%] | | | |
| | | 0.053950 | 0 | 3 | 5 | 0.0000 | | | |
| | | BEST | | | MEAN | | | nrs | fes |
| CHT | nn | CONFLICT | nac | cv | CONFLICT | nac | cv | | |
| PF | N/A | FAIL | | | | | | | |
| PFPR | 2 | 0.362361 | 2 | 7.55E-06 | 0.899420 | 2.37 | 3.74E-04 | 30 | 3.40E+05 |
| | 10 | 0.201413 | 2 | 4.35E-06 | 0.963433 | 1.67 | 1.35E-06 | 30 | 3.40E+05 |
| | 39 | 0.317522 | 0 | 2.90E-12 | 1.315517 | 0.00 | 2.61E-12 | 30 | 3.40E+05 |
| PFPPR | 2 | 0.086370 | 0 | 2.88E-12 | 0.642877 | 0.00 | 2.28E-12 | 30 | 3.40E+05 |
| | 10 | 0.314478 | 0 | 1.95E-12 | 0.719692 | 0.00 | 1.71E-12 | 30 | 3.40E+05 |
| | 39 | 0.184979 | 0 | 1.37E-12 | 0.863103 | 0.00 | 1.87E-12 | 30 | 3.40E+05 |
| PFPR+REC | 2 | **0.053983** | 3 | **1.88E-05** | **0.106360** | 2.53 | **9.40E-04** | 30 | 3.40E+05 |
| | 10 | **0.053950** | 3 | **8.84E-11** | **0.118652** | 2.37 | **4.82E-07** | 30 | 3.40E+05 |
| | 39 | **0.053965** | 0 | **3.00E-12** | **0.246805** | 0.00 | **2.97E-12** | 30 | 3.40E+05 |
| PFPPR+REC | 2 | **0.054013** | 0 | **2.55E-12** | **0.056761** | 1.57 | **6.90E-11** | 30 | 3.40E+05 |
| | 10 | **0.053964** | 0 | **3.00E-12** | **0.201187** | 0.00 | **2.91E-12** | 30 | 3.40E+05 |
| | 39 | **0.053953** | 0 | **2.99E-12** | **0.220923** | 0.00 | **2.97E-12** | 30 | 3.40E+05 |
| APM | 2 | 0.065191 | 3 | 1.24E-04 | 0.967840 | 3.00 | 4.44E-04 | 30 | 3.40E+05 |
| | 10 | 0.155968 | 3 | 1.42E-07 | 0.827705 | 3.00 | 6.38E-06 | 30 | 3.40E+05 |
| | 39 | 0.057894 | 3 | 3.22E-08 | 1.275142 | 3.00 | 7.36E-07 | 30 | 3.40E+05 |
| BM | N/A | FAIL | | | | | | | |
| BMEM | N/A | FAIL | | | | | | | |
| BMPEM | N/A | FAIL | | | | | | | |
| Toscano Pulido et al.[26] (global PSO) | | 0.068665 | - | - | 1.716426 | - | - | 30 | 3.40E+05 |
| Landa Becerra et al.[35] (CDE) | | **0.056180** | - | - | **0.288324** | - | - | 30 | 1.00E+05 |
| Muñoz Zavala et al.[34] (PESO) | | 0.081498 | - | - | 0.626881 | - | - | 30 | 3.50E+05 |
| Zheng et al.[39] (IPSO) | | 0.170852 | - | - | 0.825745 | - | - | 30 | 3.40E+05 |
| Takahama et al.[36] (αGA) | | **0.05396** | - | - | **0.19049** | - | - | 100 | 3.50E+05 |
| Takahama et al.[37] (αNSM) | | **0.053942** | - | - | **0.066770** | - | - | 30 | 7.35E+04 |
| Runarsson et al.[40] (ES + SR) | | **0.053957** | - | - | **0.067543** | - | - | 30 | 3.50E+05 |

**Table 18.** Best and mean conflicts found by our general-purpose PSO algorithm equipped with different CHTs, and those reported by other authors for reference, for the Benchmark Problem 13 (g13) as formulated in Toscano Pulido et al.[26].

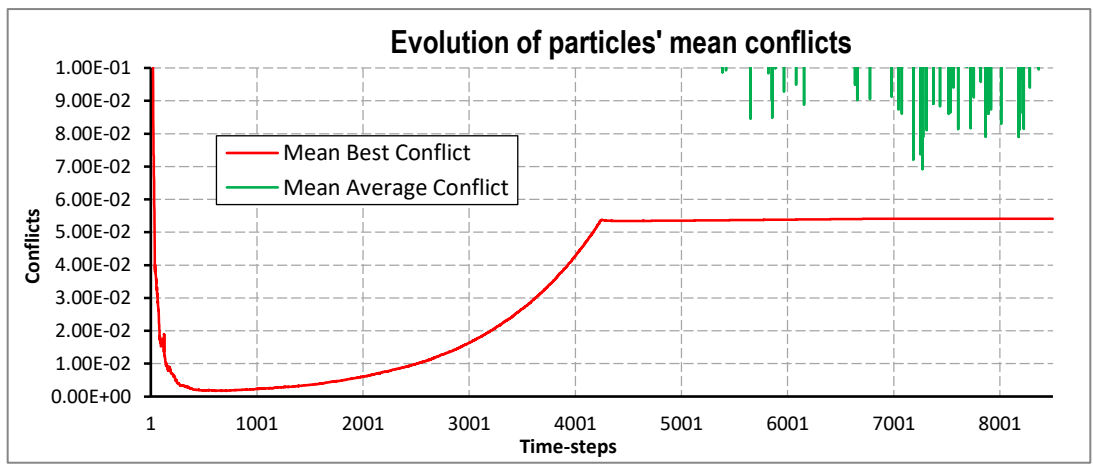

**Figure 3.** Evolution of the mean best and mean average conflicts for a PSO algorithm with REC optimizing the Benchmark Problem 13 (3 equality constraints; feasibility ratio < 0.0001%). The mean is calculated out of 30 runs, whereas the average is calculated among all particles in the swarm.

Following the same criteria as with equality constraints, the tolerance for inequality constraints will be dynamically relaxed. We expect this to be beneficial in problems with low feasibility ratio and in problems where the solution lies on or near the boundaries. We are currently working on general ways of computing the initial tolerance relaxations, and on their decrease rules.

As mentioned before, the penalization method implemented here needs to be refined, and some adaptive scheme for the penalization coefficients should be tested. Besides, the strategy of relaxing the tolerances for the constraints' violations might also be convenient, so that very high penalizations are possible without losing the ability to explore infeasible space. Although the plain bisection method is ruled out here as a general-purpose CHT, improvements are possible. A slower decrease of the momentum can be considered rather than splitting the velocities in half, while the pitfall of requiring initial feasible swarm could be tackled using priority rules and dynamically relaxed constraints.

Another improvement that will be studied consists of incorporating some diversity operator, very much like the gust of wind in Heppner and Grenander's simulations[10]; the craziness operator in initial simulations by Kennedy and Eberhart[8]; the turbulence operator in Toscano Pulido and Coello Coello's PSO algorithm[26]; or some of the mutation operators imported from EAs (e.g. Muñoz Zavala et al.'s PESO[34]). We already tried a few diversity operators, observing that they are not as straightforward as it may seem, usually alternating beneficial and harmful effects for different problems.